\documentclass{article}

\usepackage{arxiv}

\usepackage[utf8]{inputenc} 
\usepackage[T1]{fontenc}    
\usepackage{hyperref}       
\usepackage{url}            
\usepackage{booktabs}       
\usepackage{amsfonts,amsmath,amssymb}       
\usepackage{nicefrac}       
\usepackage{microtype}      
\usepackage{graphicx}
\usepackage{caption}

\usepackage{color, colortbl}

\usepackage{mathptmx,amsthm}
\usepackage{xcolor}
\usepackage{multirow}
\usepackage{subcaption}
\usepackage{tabularx,adjustbox}

\usepackage{enumitem,makecell}

\newcommand{\etal}{\textit{et al.}}
\newcommand{\ie}{\textit{i.e.,}}

\newcommand{\commenttxt}[1]{}
\newcommand{\mysubheading}[1]{\vspace{0.25cm}\noindent\textbf{#1:}}
\newcommand{\mybar}{\kern1pt\rule[-\dp\strutbox]{.8pt}{\baselineskip}\kern1pt}

\renewcommand{\mytagline}{} 

\theoremstyle{definition}

\title{Role-based Effectiveness of an Epistemic Uncertainty Estimate in Training Models for Semantic Segmentation of 3D LiDAR Point Clouds}
\author{
  Krishnakumar N., 
  Jaya~Sreevalsan-Nair\thanks{\texttt{jnair@iiitb.ac.in}} 
  \\
  Graphics-Visualization-Computing Lab,\\
  International Institute of Information Technology Bangalore, Karnataka 560100, India. \\
  \texttt{http://www.iiitb.ac.in/gvcl} \\
}
  
\begin{document}

\maketitle

\begin{abstract}
Point cloud segmentation is a fundamental task in 3D data analysis, with applications across robotics, augmented and virtual reality, and remote sensing. However, the inherent sparsity, noise, and unordered nature of point clouds introduce significant challenges, particularly in achieving reliable classification. While the state-of-the-art rely on training models for explicit modeling of uncertainty, there is a gap in exploring different roles that estimated uncertainty using generative approaches can play. This work investigates the impact of deterministic uncertainty in point cloud segmentation by explicitly modeling epistemic uncertainty and using it in different roles, namely, as a feature and in loss function. We evaluate a baseline artificial neural network (ANN) and two uncertainty-aware variants (uncertainty as an additional input feature and as a per-sample confidence weight in the training loss) on four large-scale LiDAR benchmarks spanning airborne and mobile acquisition, namely ISPRS Vaihingen, AHN3, Paris-Lille-3D, and Toronto-3D, together with PointNet++ and RandLA-Net references and their uncertainty-augmented counterparts under canonical spatially disjoint train and test splits. Under this protocol, uncertainty-aware modeling improves mean Intersection over Union, recovers minority classes that the baseline ANN never predicts at all, and in half of the configurations improves the alignment between prediction error and estimated uncertainty as measured by Patch Accuracy versus Patch Uncertainty. Our findings indicate that uncertainty can play multiple roles in the segmentation pipeline rather than as a device for improving accuracy alone.
\end{abstract}


\section{Introduction} \label{sec:introduction}
The increased popularity of Light Detection and Ranging (LiDAR) as a 3D sensing technologies has led to a rapid increase in the availability of point cloud data, enabling transformative applications in autonomous navigation, urban mapping, robotics, and immersive environments such as augmented and virtual reality. A central task in these domains is accurate point cloud segmentation, which involves assigning semantic labels to individual 3D points. Unlike structured image data, point clouds are unordered, irregular, and often contaminated with noise, which makes segmentation inherently difficult and makes the resulting predictions harder to trust. Very few semantic segmentation models currently incorporate uncertainty from the data~\cite{shojaei2025uncertainty}.

LiDAR point clouds can be acquired from different platforms, notably airborne laser scanning (ALS)~\cite{rottensteiner2012isprs,ahn3,varney2020dales,ye2020lasdu}, which captures scenes primarily from above, and mobile laser scanning (MLS)~\cite{tan2020toronto,roynard2018paris}, which observes them from ground-level trajectories; these differing viewpoints lead to distinct coverage patterns, occlusions, and point-density distributions~\cite{puente2013mobile}.

\begin{figure*}[h]
    \centering
    \includegraphics[width=\textwidth]{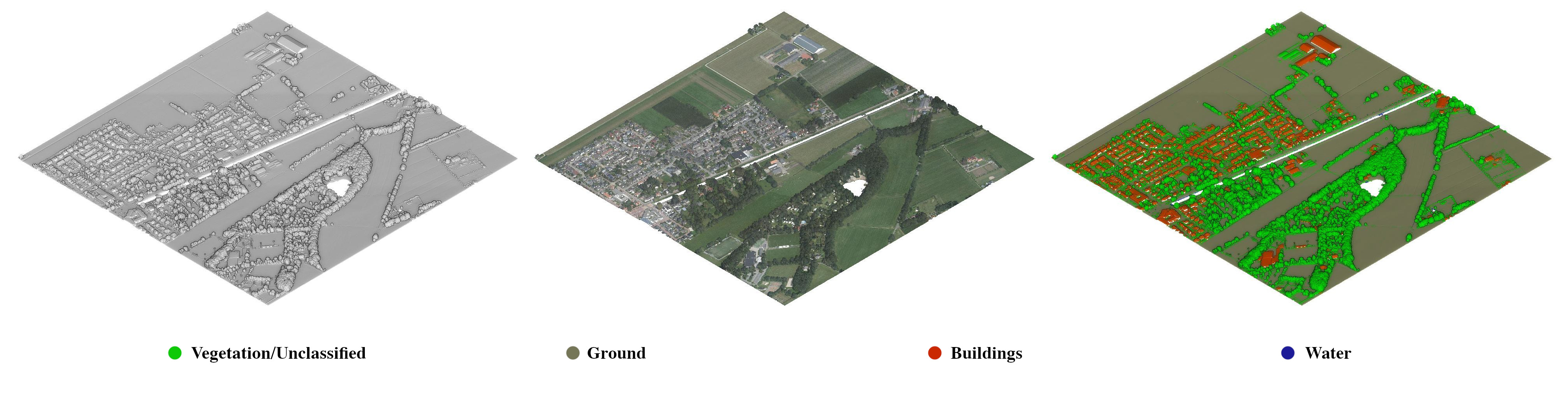}
    \caption{Depiction of point cloud segmentation, comparing raw point clouds, RGB data, and segmented outputs (28AN1\_16 tile, Lemelerveld region from the AHN3 dataset~\cite{ahn3}). The legend shows the classes present in the tile; class 0 (never classified) and class 26 (reserved) are excluded.}
    \label{fig:teaser}
\end{figure*}

Traditional deep learning methods for point cloud segmentation primarily focus on maximizing accuracy but often overlook uncertainty quantification, which is critical for reliable decision making in safety-sensitive applications. Uncertainty in this context arises from two primary sources --- \textit{epistemic} uncertainty reflects model limitations due to finite training data~\cite{kendall2017uncertainties}, while \textit{aleatoric} uncertainty captures noise and variability inherent in the sensor measurements~\cite{hullermeier2021aleatoric}. Their inherent characterization makes aleatoric irreducible and epistemic reducible~\cite{postels2022practicality}. \textit{Total} uncertainty is the sum of aleatoric and epistemic uncertainty. Epistemic uncertainty is the component involved in spatial generalization, and hence, it is of interest in this work. Properly modeling these uncertainties not only improves prediction robustness but also provides confidence estimates that are essential for downstream systems, such as autonomous robots or remote sensing platforms, where erroneous classifications can have severe consequences.

Despite substantial progress in point cloud segmentation, most existing approaches focus primarily on maximizing classification accuracy, while uncertainty estimation is often treated as a secondary or implicit byproduct. In safety-critical applications such as autonomous navigation, urban mapping, and robotic perception, understanding where and why a model is uncertain is often as important as the prediction itself~\cite{araujo2024roadsafety}. Uncertainty estimates enable downstream systems to reason about prediction reliability, reject unreliable outputs, and prioritize human intervention or additional sensing. However, there is a significant gap in exploring the use of estimated uncertainty values in a deep learning model for semantic segmentation of ALS and TLS point clouds, other than providing the values to improve the trustworthiness of the point-wise classification.

This is not a proposal for a new segmentation architecture, instead an empirical study of uncertainty behavior in a segmentation pipeline. We deliberately keep the core predictive model simple, a compact point-wise multilayer perceptron over handcrafted geometric descriptors, so that changes in behavior can be attributed to the uncertainty treatment rather than to architectural capacity.

The primary objective is to compute uncertainty, analyze its impact on segmentation reliability, and visualize its spatial distribution across different semantic classes and datasets. All experiments use canonical, spatially disjoint train and test splits, so that the test region is never seen during training or during uncertainty estimation~\cite{roberts2017cv,ploton2020}. We regard this as a prerequisite for any meaningful claim about uncertainty-aware modeling, since the value of an epistemic uncertainty estimate can only be assessed when the model is genuinely asked to extrapolate.

We propose to systematically study the impact of uncertainty on point cloud segmentation by exploring its role as a feature and in the loss function. We evaluate three neural network (NN)-based approaches:
\begin{enumerate}
    \item A baseline segmentation model trained on extracted point cloud features.
    \item A model that incorporates epistemic uncertainty as an explicit feature to augment classification.
    \item A model trained with a custom confidence-weighted loss function designed to penalize uncertain samples during optimization.
\end{enumerate}

We additionally include PointNet++~\cite{qi2017pointnetpp} and RandLA-Net~\cite{hu2019randla}, together with an uncertainty-augmented variant of each, as references, so that the reported numbers can be situated against widely used point based architectures rather than being read in isolation. Both architectures are also considerably stronger than the point-wise network on these benchmarks, which allows us to examine whether the effect of the uncertainty treatment persists as the model becomes stronger.

\clearpage
Our contributions are:
\begin{itemize}
    \item Adapting a point-wise epistemic uncertainty estimate obtained from a class-conditional Gaussian density in the geometric feature space, fitted on the training split alone. This estimate used for out-of-distribution (OOD) detection in vehicle LiDAR semantic segmentation~\cite{shojaei2025uncertainty} is re-purposed in this work as a proxy of epistemic uncertainty~\cite{postels2022practicality}. Because the estimate does not depend on the trained network, the same uncertainty map is shared by all model variants, which isolates the effect of how uncertainty is used from the effect of how well it is estimated.
    \item A systematic evaluation of the role of the epistemic uncertainty estimate, as a feature and loss function, across four benchmarks spanning airborne and mobile acquisition, under canonical spatially disjoint splits.
    \item Modified PointNet++ and RandLA-Net implementations as the uncertainty-augmented counterparts of two widely used point-based architectures for the evaluation.
    \item A reliability-centered analysis using per-class IoU and PAvPU, which shows that the practical benefit of uncertainty-aware modeling lies in minority class recovery and in error and uncertainty alignment, and that the mIoU gain persists even for the strongest architecture.
\end{itemize}

 The code for this work is available at:\\ \url{https://github.com/aiatoz/Uncertainty-Aware-LiDAR-Semantic-Segmentation}

\section{Related Work} \label{sec:relatedwork}
Uncertainty estimation has become an important part of 3D point cloud learning, particularly where segmentation or detection models need to distinguish reliable predictions from those made in unfamiliar or ambiguous regions. Early work mainly focused on Bayesian approaches that explicitly model uncertainty. One pioneering study~\cite{petschnigg2021uncertainty} uses a Bayesian neural network with dropout-based sampling for point cloud segmentation and shows that the resulting uncertainty can improve the reliability of 3D modelling. Such sampling-based estimates are expensive at the scale of a large point cloud, since Monte Carlo dropout requires a separate stochastic forward pass for every sample drawn. Two lines of work address this cost. Firstly, an efficient variant~\cite{qi2023nsa} replaces repeated forward passes with a single one, by observing that the stochastic inferences of a point's spatial neighbors can approximate the repeated stochastic inferences of the point itself, so that a point-wise predictive distribution is established by aggregating the probabilistic outputs of its neighborhood. Secondly, a lightweight post-processing framework~\cite{riedlinger2025} estimates uncertainty and prediction quality for LiDAR object detection without modifying the architecture or the training procedure, thereby avoiding prediction sampling altogether. These approaches illustrate the trade-off between richer uncertainty estimates and the computational cost of obtaining them, and the estimate in each case remains tied to the trained network and to its architecture.

Eigenvalue-based descriptors over local neighborhoods are a long standing basis for point cloud interpretation. One comprehensive study~\cite{weinmann2015semantic} analyzes how neighborhood size, feature relevance and classifier choice interact, and shows that selecting the neighborhood per point rather than globally materially affects classification performance~\cite{weinmann2014semantic}. In contrast to these handcrafted geometric descriptors, recent approaches have explored learned local feature representations to improve large-scale point cloud segmentation. For example, Chen et al. introduced LFE-Net, which combines local feature enhancement with simple offset attention and incorporates class-imbalance-aware sampling to improve segmentation performance~\cite{chen2024lfenet}.

Learned point-based architectures form the other main basis for point cloud segmentation~\cite{wu2024point}~\cite{lai2022stratified}~\cite{qian2022pointnext}. One pioneering hierarchical network, PointNet++~\cite{qi2017pointnetpp}, applies a shared point-wise network recursively over nested local regions selected by farthest point sampling, so that local structure is captured at increasing scales. Its sampling and grouping steps, however, become expensive on large scenes. RandLA-Net~\cite{hu2019randla, hu2021learning} addresses this by replacing farthest point sampling with random sampling, and compensates for the information lost during sampling through a local feature aggregation module that progressively increases the receptive field of each point, which allows large scale point clouds to be processed directly. We use both architectures as references and supply the same uncertainty estimate to each as an additional input feature.

We can estimate uncertainty directly from the density of the learned representation. One prominent approach~\cite{lee2018mahalanobis} models class-conditional feature distributions with Gaussian discriminant analysis and uses the resulting Mahalanobis distance as a measure of how far a sample lies from the training distribution. A recent study~\cite{mukhoti2023ddu} extends this idea through Deep Deterministic Uncertainty (DDU), showing that a Gaussian mixture model (GMM) fitted to the features of a deterministic network can provide a competitive uncertainty estimate without Bayesian sampling. Similar to DDU, other deterministic uncertainty methods (DUMs)~\cite{postels2022practicality}, such as Directional Concentration Uncertainty (DCU)~\cite{winkens2020contrastive} and Maximally Informative Representation(MIR)~\cite{postels2021hidden}, use the class-conditional GMMs as a proxy for deterministic epistemic uncertainty. 

A related formulation has been applied to point clouds in~\cite{yao2024uncertainty}, where a prototype entropy metric measures the entropy of the cosine similarities between a point embedding and a set of class prototypes. The motivation given there is that softmax confidence is an unreliable indicator of correctness, since networks can be both overconfident and underconfident, so that predictions which are confidently wrong are admitted by confidence thresholding. Prototype entropy uses only the class means, whereas a full class conditional Gaussian additionally models the covariance of each class, so that displacement along a direction in which a class naturally varies is treated differently from displacement along a direction in which it does not.

Our approach also formulates the deterministic epistemic uncertainty using GMMs as a proxy. However, two properties distinguish it from the state of the art. First, the descriptors are handcrafted rather than learned by the segmentation network, so the estimate is independent of the model being evaluated and the same uncertainty map can be supplied to different architectures. Second, the estimate is obtained before training rather than during it, since it requires no embeddings, no prototypes and no iterative updates.

More recently, a closely related work~\cite{shojaei2025uncertainty} applies this idea to LiDAR scene segmentation, fitting class-conditional GMMs in the feature space of a discriminative network to obtain epistemic uncertainty and out of distribution scores. An alternative framework~\cite{xu2025propagation} takes a related view from the geodetic side, predicting point level uncertainty in mobile laser scanning data directly from geometric descriptors, and reports that point cloud uncertainty is learnable from local geometry alone.

Uncertainty has also been incorporated directly into 3D segmentation and learning frameworks. One notable framework~\cite{yang2023geometry} combines geometry aware learning with uncertainty aware pseudo labelling for class incremental 3D semantic segmentation, using uncertainty to filter unreliable pseudo labels and reduce the effect of semantic shift. 

A related framework~\cite{zhou2025dcuf} represents the reliability of unlabelled points by combining the confidence of the predicted distribution with its Shannon entropy~\cite{shannon1948math}, and uses the result as an uncertainty-aware loss over all unlabelled points rather than discarding those below a confidence threshold.

Another recent method~\cite{mei2024improving} uses local neighborhood information and class reliability together with Markov Random Field (MRF)-based refinement to improve the spatial consistency of point cloud predictions. These approaches use uncertainty together with additional spatial or learning mechanisms, whereas our formulation uses geometric feature density to obtain the uncertainty estimate independently of the segmentation model.

Uncertainty is also widely used for active learning, where the goal is to identify the points that are most useful to annotate. One multi-scale approach~\cite{xu2023hierarchical} proposes a hierarchical point-based active learning framework that estimates uncertainty at multiple spatial scales and combines it with feature distance to select uncertain but representative points. A subsequent study~\cite{li2025llm} further explores hierarchical uncertainty for 3D point cloud active learning, using an LLM generated semantic taxonomy to guide point selection.

Evaluating an uncertainty estimate requires more than measuring segmentation accuracy alone. A widely used metric~\cite{mukhoti2018pavpu} introduced Patch Accuracy versus Patch Uncertainty (PAvPU) to measure whether a model tends to be confident when it is correct and uncertain when it is wrong. We adopt this idea and extend it from image patches to local 3D neighborhoods in Section~\ref{sec:evalprotocol}.

Taken together, existing work shows that uncertainty can be estimated through Bayesian sampling, learned feature distributions, or task-specific prediction models, and can be used for prediction refinement, incremental learning and active data selection. Our work focuses on a simpler setting in which uncertainty is obtained directly from local geometric descriptors and kept independent of the segmentation model. This allows the same uncertainty estimate to be incorporated into different models for an empirical study to determine whether geometric uncertainty itself is related to segmentation errors under a spatially disjoint evaluation.

\begin{table*}[t]
\centering
\caption{Geometric descriptors computed for each point from the eigenvalues and eigenvectors of the local covariance matrix}
\label{tab:features}
\renewcommand{\arraystretch}{1.5}
\setlength{\tabcolsep}{8pt}
\begin{tabular}{@{}l c p{0.5\linewidth}@{}}
\toprule
Descriptor & Definition & Interpretation \\
\midrule

Eigenvalue sum & $\Sigma_\lambda = \lambda_1 + \lambda_2 + \lambda_3$ & Total variance of the neighborhood, independent of how it is distributed among directions. Larger values indicate greater overall spatial extent. \\

PCA1 & $\mathrm{PCA}_1 = \dfrac{\lambda_1}{\Sigma_\lambda}$ & Proportion of total variance along the dominant axis. \\

PCA2 & $\mathrm{PCA}_2 = \dfrac{\lambda_2}{\Sigma_\lambda}$ & Proportion of total variance along the second axis. \\

Linearity & $L = \dfrac{\lambda_1 - \lambda_2}{\lambda_1}$ & How well the neighborhood aligns along a dominant axis. Values close to 1 indicate linear structure such as edges, wires or poles. \\

Planarity & $P = \dfrac{\lambda_2 - \lambda_3}{\lambda_1}$ & How planar the neighborhood is. High values indicate that most points lie on a surface such as a wall, roof or the ground. \\

Sphericity & $S = \dfrac{\lambda_3}{\lambda_1}$ & Isotropy of the local point distribution. Larger values indicate a volumetric neighborhood with no dominant direction, as in clutter or foliage. \\

Surface variation & $SV = \dfrac{\lambda_3}{\Sigma_\lambda}$ & Proportion of variance along the least varying direction. Higher values indicate a locally irregular or curved surface. Unlike sphericity, the normalization is by the total variance rather than by $\lambda_1$. \\

Anisotropy & $A = \dfrac{\lambda_1 - \lambda_3}{\lambda_1}$ & Contrast between the strongest and weakest variance directions. Higher values indicate elongated or planar structure rather than isotropic shape. \\

Omnivariance & $O = (\lambda_1 \lambda_2 \lambda_3)^{1/3}$ & Geometric dispersion in all directions, measuring how spread out the neighborhood is in 3D space. \\

Eigenentropy & $\tilde{\lambda}_i = \dfrac{\lambda_i}{\Sigma_\lambda}, \quad  E = -\sum_{i=1}^{3} \tilde{\lambda}_i \log (\tilde{\lambda}_i)$ & Shannon entropy~\cite{shannon1948math} of the distribution formed by the normalized eigenvalues, quantifying the disorder of the local shape. Higher entropy indicates a more complex and less regular structure. \\

Verticality & $V = 1 - |e_{3,z}|$ & Orientation of the local surface normal relative to the vertical axis. Approaches 0 for horizontal surfaces such as the ground and 1 for vertical surfaces such as walls. \\

Normal components & $(n_x, n_y, n_z)$ & Components of the estimated local surface normal $e_3$, the eigenvector associated with $\lambda_3$, giving the orientation of the neighborhood in 3D. The sign is not resolved, so verticality uses $|e_{3,z}|$. \\

\bottomrule
\end{tabular}
\end{table*}

\section{Background} \label{sec:features}
Robust feature computation is critical for effective classification of 3D points. Initial attempts involved using CloudCompare and custom PySpark implementations. However, these either produced unreliable features or suffered from unusually long computation times when scaled to large datasets. To address these limitations, the \texttt{Jakteristics}~\cite{jakteristics} Python library was adopted, which efficiently computes geometric and structural descriptors while maintaining accuracy comparable to manually computed features. 

To characterize local 3D shape properties, handcrafted geometric features are computed from the eigenvalues (Weinmann \etal~\cite{weinmann2015semantic}) of the covariance matrix of each point's neighborhood. Given a query point $X_0$, let $\mathcal{N}(X_0)$ denote its set of neighboring points within a fixed search radius. The local covariance matrix is defined as

\begin{equation}
C(X_0) = \frac{1}{|\mathcal{N}(X_0)|} \sum_{X_i \in \mathcal{N}(X_0)} (X_i - \bar{X})(X_i - \bar{X})^{\top},
\end{equation}
where $\bar{X}$ is the centroid of the neighbors. Let $\lambda_1 \geq \lambda_2 \geq \lambda_3 \geq 0$ be the ordered eigenvalues of $C(X_0)$, and let $e_1, e_2, e_3$ be the corresponding eigenvectors, with $e_3$ giving the estimated local surface normal direction. The eleven eigenvalue-based descriptors computed from these quantities are listed in Table~\ref{tab:features}, together with the three components of the surface normal $(n_x, n_y, n_z)$ they are combined with.

\section{Methodology} \label{sec:methodology}

\begin{figure*}[t]
\centering
\includegraphics[width=\textwidth]{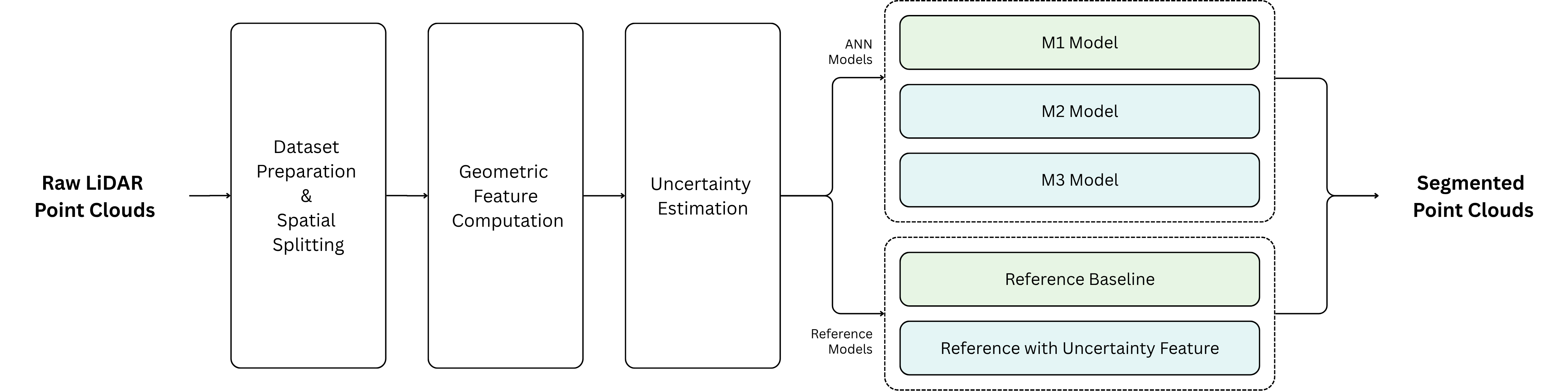}
\caption{Overview of the uncertainty-aware segmentation pipeline. Raw LiDAR point clouds undergo dataset preparation and spatial splitting, geometric feature computation, and epistemic uncertainty estimation (Section~\ref{sec:uncertainty}); the last of these is computed once and shared, unchanged, across every downstream model. The resulting features and uncertainty estimates are then used to train two parallel model families: the three feed-forward ANN variants M1, M2, and M3 (Section~\ref{sec:models}), and the reference architectures, PointNet++ and RandLA-Net, each trained as a baseline and as an uncertainty-augmented counterpart, yielding the final segmented point clouds.}
\label{fig:pipeline}
\end{figure*}

The proposed methodology is structured into five stages, executed in a fixed order so that no statistic computed at a later stage can influence an earlier one. These are dataset preparation and spatial splitting, geometric feature computation, uncertainty estimation, classification with explicit uncertainty integration, and evaluation. Each stage writes a single artefact that the next stage consumes, so that every number reported for a given dataset refers to exactly the same partition of points.
    
\subsection{Dataset Preparation and Spatial Splitting}
\label{sec:splitting}

For the experiments, outdoor large-scale LiDAR point cloud datasets are employed, where each point is represented by three spatial coordinates $(x,y,z)$ along with an associated class label. The set of class labels varies across datasets, but in all cases every point is annotated as belonging to a specific class.

To prepare the data for feature computation, the raw files are restructured into a tabular format. This preprocessing step splits the raw record into individual attributes, discards columns that are not used, such as intensity, RGB, GPS time and scan angle rank, and converts the remaining values into numeric form. Parsing validates the observed column count against a per-dataset specification and the label set against the declared class map, to prevent silently misaligned rows and values propagating to the later stages.

Train and test partitioning is file-based and spatial. For each dataset we adopt a canonical partition in which entire tiles or scan sections are assigned to either the training or test side, following the dataset authors' convention where one exists (except for Paris-Lille-3D, since it follows a separate test split and a submission-based evaluation protocol). The partitions used are listed in Table~\ref{tab:TrainTestSplit}. Under this file-level split, the test region is spatially disjoint. The test region might be acquired in a separate pass and may differ in point density, class balance and scene composition. This is a considerably harder evaluation than a random point-wise split.

Feature scaling is fitted on the training split only and applied unchanged to the test split, so that the value ranges of the held out region do not influence the scaler. Prior to scaling, the coordinates $(x,y,z)$ are re-centred per source file relative to its own centroid. This re-centring is a pure geometric translation that uses only that file's own coordinates, borrowing no statistic from another file and no information from labels, and it is therefore correct to apply to unseen test tiles. Its purpose is to prevent absolute geospatial position from becoming a learnable signal. Without it, a model can partially memorise where a point is rather than what its neighborhood looks like, and that behavior does not transfer to a new tile.

In summary, we avoided data leakage between the training and test splits and ensured that the model is evaluated on completely unseen spatial regions. For spatial datasets, the IID assumption is generally difficult to satisfy because nearby observations are often spatially autocorrelated, with nearby points tending to exhibit similar characteristics~\cite{roberts2017cv}. For this reason, we consider spatially disjoint train-test splitting to be a crucial part of evaluating generalization in large-scale LiDAR point clouds.

\begin{table}[t]
\centering
\caption{Canonical file level splits used for all experiments. The test region is disjoint in space from the training region and is never used for training, for fitting the feature scaler, or for fitting the uncertainty model.}
\label{tab:TrainTestSplit}
\begin{tabular}{llll}
\toprule
Dataset & Modality & Training files & Test file \\
\midrule

Vaihingen & ALS & Area 2 & Area 3 \\

\addlinespace
AHN3 & ALS & \begin{tabular}[t]{@{}l@{}}28AN1\_16 \\ 28AZ1\_13\end{tabular} & 32CN2\_11 \\

\addlinespace
Paris-Lille-3D & MLS & \begin{tabular}[t]{@{}l@{}}Lille1\_1 \\ Lille1\_2 \\ Lille2\end{tabular} & Paris \\

\addlinespace
Toronto-3D & MLS & \begin{tabular}[t]{@{}l@{}}L001 \\ L002 \\ L004\end{tabular} & L003 \\

\bottomrule
\end{tabular}
\end{table}

\subsection{Geometric Feature Computation}
\label{sec:featurecomputation}

Using the geometric descriptors defined in Section~\ref{sec:features}, we compute a per-point feature representation as follows.

In total, fourteen descriptors are computed per point, namely eigenvalue sum, omnivariance, eigenentropy, anisotropy, planarity, linearity, the first two principal component projections PCA1 and PCA2, surface variation, sphericity, verticality, and the three normal vector components $(n_x, n_y, n_z)$. These are merged with the re-centred and scaled coordinates $(x,y,z)$ to give a 17 dimensional feature vector per point, which is the input to the ANN models.

Points whose neighborhood within the search radius is too sparse to yield a stable eigen-decomposition are returned as undefined by the feature extractor. Rather than dropping these silently, they are flagged explicitly at extraction time and counted per class, so that any density correlated bias introduced by their removal remains visible. This matters specifically for AHN3, as discussed in Section~\ref{sec:ahn3}.

A search radius of $0.25$ m is used throughout, selected from a sweep over $\{0.10, 0.15, 0.20, 0.25\}$ m~\cite{weinmann2015semantic} on the training splits. The transfer of a single radius across acquisition platforms of very different point density is noted as a limitation in Section~\ref{sec:limitations}.

\subsection{Deterministic Epistemic Uncertainty Estimation}
\label{sec:uncertainty}
This work quantifies the deterministic epistemic uncertainty, which reflects model uncertainty due to limited data representation and capacity. We model each class's distribution in the geometric feature space using a multivariate Gaussian, following the Gaussian Discriminant Analysis (GDA) formulation in its quadratic form, that is, with a class-specific covariance rather than a shared one.

Let $x_i \in \mathbb{R}^{d}$ denote the $d = 14$ dimensional vector of geometric descriptors for point $i$. The coordinates $(x,y,z)$ are deliberately excluded from $x_i$ here, for reasons given below. For each class $c \in \{1, \dots, C\}$, we estimate the class-conditional probability, from the training split alone:
\begin{equation}
p(x_i \mid c) = \mathcal{N}(x_i; \mu_c, \Sigma_c),
\label{eq:gaussian}
\end{equation}

where $\mu_c$ and $\Sigma_c$ are the empirical mean and covariance of the training points of class $c$. Covariances are regularized as $\Sigma_c \leftarrow \Sigma_c + \epsilon I$ with $\epsilon = 10^{-3}$ to improve numerical stability, particularly for minority classes, such as bollards, trash cans and pedestrians in Paris-Lille-3D, and utility lines and poles in Toronto-3D, whose sample counts are close to the feature dimensionality. The resulting regularized covariance matrices are used to compute the log-determinant and inverse terms in the Gaussian likelihood.

The point-wise epistemic uncertainty $u_i$ is defined as the negative log-likelihood of $x_i$ under its highest class-conditional likelihood:
\begin{equation}
u_i = - \max_{c \in \{1,\dots,C\}} \log \mathcal{N}(x_i; \mu_c, \Sigma_c).
\label{eq:uncertainty}
\end{equation}
Higher values of $u_i$ indicate that the point lies far from any well supported region of the feature space, and hence that it is more uncertain. Equation~\ref{eq:uncertainty} gives the \textit{hard assignment} of epistemic uncertainty when using the $\max$ operator on the class-wise conditional probabilities, which is in contrast to the \textit{soft assignment}. Soft assignment is computed as the negative marginal log-likelihood, \ie~using the natural logarithm of the class-wise conditional probabilities. Hard assignment gives non-overlapping clusters needed for semantic segmentation, unlike the overlapping clusters from soft assignment, which is used for ambiguous labeling in expectation-maximization training.

Expanding Equation~\ref{eq:uncertainty}, $u_i$ is the minimum, over classes, of the sum of a squared Mahalanobis distance and a log-determinant term, up to a constant. It therefore provides a class-conditional feature-space density score~\cite{lee2018mahalanobis, mukhoti2023ddu}. Class priors are deliberately omitted from Equation~\ref{eq:uncertainty}, so that the resulting quantity reflects geometric typicality rather than class frequency. Min-max normalized likelihood is given by $\tilde{u}_i$:
\begin{equation}
\tilde{u}_i = \frac{u_i - \min_{j \in \mathcal{T}} u_j}
{\max_{j \in \mathcal{T}} u_j - \min_{j \in \mathcal{T}} u_j},
\label{eq:normalized}
\end{equation}
where $\mathcal{T}$ denotes the training split. The same extrema are applied
unchanged to the test split, so $\tilde{u}_i$ is bounded in $[0,1]$ on training
data but may exceed 1 on test data. Test uncertainties are consequently permitted to exceed 1. This is intentional, since a test tile containing geometry outside the training distribution should be able to register as such rather than being renormalized back into range.

Three properties of this estimator determine how the results should be read, and we state them explicitly.

\mysubheading{The estimate is model independent}
Equation~\ref{eq:uncertainty} depends only on the training features and labels, and not on any trained network. The same uncertainty map is therefore shared by all model variants, the baseline ANN and both uncertainty-aware ANN variants as well as the PointNet++ and RandLA-Net variants. This is a design choice. It means that any difference in PAvPU between model variants reflects a difference in where those models make their errors, with the uncertainty estimate held fixed, rather than a difference in the quality of two separate uncertainty estimators.

\mysubheading{The estimate is a density-based proxy, not a Bayesian posterior}
We do not claim a posterior over network weights. What Equation~\ref{eq:uncertainty} measures is how atypical a point's local geometry is relative to the geometry covered by the training data, which is the operational content of epistemic uncertainty in a spatial transfer setting, and which is precisely what a file level split stresses. The term epistemic is used in this sense throughout, consistent with the deterministic density-based literature~\cite{lee2018mahalanobis, mukhoti2023ddu}.

\mysubheading{Coordinates are excluded from the density model}
Fitting Equation~\ref{eq:gaussian} on a feature vector that includes absolute $(x,y,z)$ could make points from a held-out tile appear highly uncertain because their coordinates differ from those represented in the training data. This would reflect geographic displacement rather than geometric novelty. Excluding the coordinates makes the estimate a statement about local geometric structure captured by the descriptors, which is the property intended to transfer across tiles.

\subsection{Classification and Uncertainty Integration}
\label{sec:models}

\begin{figure*}[t]
\centering
\includegraphics[width=\textwidth]{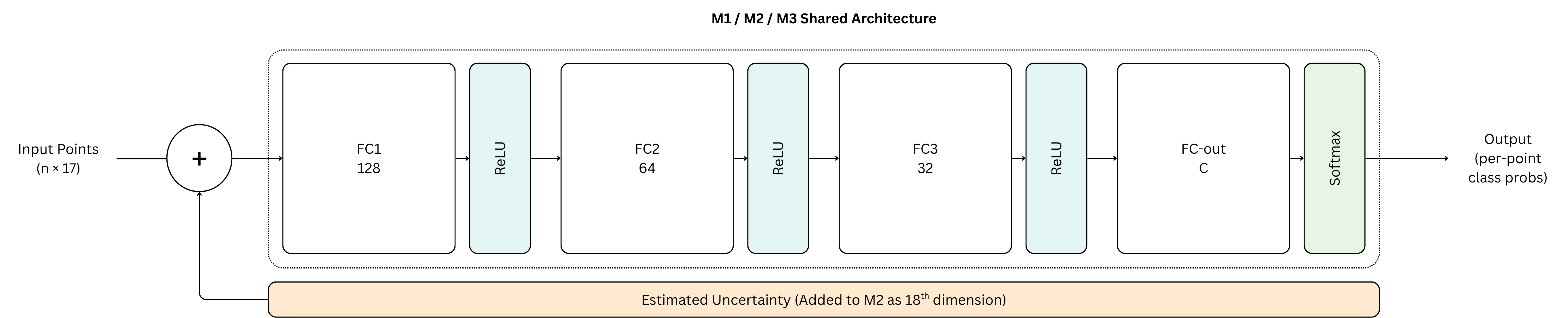}
\caption{Shared feed-forward architecture used by M1, M2, and M3 (Section~\ref{sec:models}). All three variants use an identical three-hidden-layer network (128, 64, and 32 units, each followed by ReLU) and a final $C$-unit softmax layer, where $C$ is the number of semantic classes for the dataset. M1 and M3 take the 17-dimensional feature vector directly as input; M2 additionally concatenates the normalized epistemic uncertainty $\tilde{u}_i$ (Eq.~\ref{eq:normalized}) as an 18th input dimension, shown here as the uncertainty branch merging into the network input. M3 uses the same 17-dimensional input as M1 and is not distinguished in this forward-pass diagram; it differs only in its training objective, where $\tilde{u}_i$ instead reweights the per-sample cross-entropy loss (Eqs.~\ref{eq:weight}-\ref{eq:weightedloss}).}
\label{fig:ann_architecture}
\end{figure*}

The classification framework uses neural network architectures that explicitly incorporate uncertainty estimation. All three variants use the same fully connected feed-forward network. The network takes the per-point feature vector as input and has three hidden layers with 128, 64 and 32 neurons, respectively, with ReLU activation at each layer. The final layer has one neuron per class with softmax activation, producing a probability distribution over the classes. The input and output dimensions are derived from the data rather than fixed in advance. The network is intentionally kept relatively small, since the focus of this study is the effect of uncertainty treatment rather than network capacity.

The three modeling approaches evaluated are as follows :

\begin{enumerate}
    \item \textbf{M1, Baseline neural network :} A feed-forward neural network trained on the 17 dimensional feature vector, including the re-centred and scaled coordinates together with the fourteen geometric descriptors, using unweighted categorical cross-entropy.
    
    \item \textbf{M2, Epistemic uncertainty-augmented model:} Identical to M1, except that the normalized uncertainty $\tilde{u}_i$ of Equation~\ref{eq:normalized} is appended as an eighteenth input dimension. For every point, in both splits, $\tilde{u}_i$ is obtained by evaluating that point's own geometric descriptor vector under the class-conditional Gaussians of Equation~\ref{eq:gaussian}, which are fitted on the training split alone. The maximum in Equation~\ref{eq:uncertainty} is taken over all classes, so no label is required at scoring time, and no label information from the held out tile enters the model.
    
    \item \textbf{M3, Custom loss function model:} A model trained with a confidence-weighted loss function~\cite{kendall2017uncertainties,qi2023nsa}, where points with higher confidence, and hence lower uncertainty, are assigned greater importance during optimization. This design mitigates the effect of noisy or ambiguous samples. The inputs are identical to those of M1, with no additional feature. Using the normalized uncertainty $\tilde{u}_i$ of Equation~\ref{eq:normalized}, the per-sample weight and the resulting objective are:
    
\begin{equation}
  w_i = 1 - \tilde{u}_i; \label{eq:weight}
\end{equation}

\begin{equation}
  \mathcal{L} = \frac{1}{N} \sum_{i=1}^{N} w_i \left( - \sum_{c=1}^{C} y_{i,c} \log \hat{y}_{i,c} \right), \label{eq:weightedloss}
\end{equation}

    where $N$ denotes the total number of training samples and $C$ is the number of semantic classes. The scalar weight $w_i$ controls the contribution of sample $i$ to the loss and is defined as an inverse function of uncertainty, so that $w_i = 1$ for the most confident training point and $w_i = 0$ for the least. The vector $y_i = [y_{i,1}, \dots, y_{i,C}] \in \{0,1\}^{C}$ denotes the one-hot encoded ground-truth label, and $\hat{y}_i = [\hat{y}_{i,1}, \dots, \hat{y}_{i,C}]$ is the predicted class probability vector obtained by applying a softmax activation to the network output. The term $-\sum_{c=1}^{C} y_{i,c} \log \hat{y}_{i,c}$ corresponds to the standard categorical cross-entropy loss for sample $i$. Since the loss is evaluated only during training, $\tilde{u}_i$ is bounded in $[0,1]$ throughout and $w_i$ is non-negative by construction.     
\end{enumerate}

In implementation, the per-sample weights are supplied to the training loop through the framework's native sample weighting mechanism, so that each weight remains aligned with its sample under batch shuffling across epochs.

For the point-based references, we additionally train \textbf{PointNet++}~\cite{qi2017pointnetpp} and \textbf{RandLA-Net}~\cite{hu2019randla,hu2021learning}, each in a baseline configuration and in a variant that consumes the same $\tilde{u}_i$ as an additional per-point input channel, under identical canonical splits.

\subsection{Evaluation Protocol}
\label{sec:evalprotocol}

The following commonly-used metrics are used for evaluation of the effectiveness of the modeling approaches:
\begin{itemize}

    \item \textbf{Classification Accuracy:} The percentage of correctly classified points, measuring overall performance. Because all four datasets are severely class imbalanced, accuracy is dominated by the ground, road and building classes, and we do not treat it as the primary metric.
    
    \item \textbf{Cross-Entropy Loss:} The mean per-point categorical cross-entropy between the predicted class distribution and the ground-truth label, monitored during training as the optimization objective. We report it for completeness but do not treat it as an evaluation metric, since it is also dominated by whichever classes are most frequent in a given test tile.

    \item \textbf{Mean Intersection over Union (mIoU):} A standard metric~\cite{petschnigg2021uncertainty} in segmentation pipelines, which evaluates how well a model's predicted segmentation matches the ground truth and which penalizes both false positives and false negatives. For class $i$

    \begin{equation} \mathrm{IoU}_i = \frac{TP_i}{TP_i + FP_i + FN_i}, \qquad \mathrm{mIoU} = \frac{1}{|\mathcal{C}'|} \sum_{i \in \mathcal{C}'} \mathrm{IoU}_i, \end{equation} 
    
    where $\mathcal{C}'$ is the set of classes that are both non-excluded and actually present in the test split. Classes that are declared in a dataset taxonomy but absent from the held out tile are reported explicitly and omitted from the average, rather than silently contributing a zero, since including them would make mIoU a function of which tile happened to be held out. Unclassified categories are excluded from mIoU for Paris-Lille-3D and Toronto-3D.

    \item \textbf{Patch Accuracy versus Patch Uncertainty (PAvPU):} PAvPU~\cite{mukhoti2018pavpu} measures whether uncertainty is aligned with prediction errors. Ideally, correct predictions should have low uncertainty, while incorrect predictions should have high uncertainty.
    
    For point clouds, we define a patch as the $K$ nearest neighbors of each point, with $K=30$. Patch accuracy is the fraction of correctly classified points in the patch, while patch uncertainty is the mean normalized uncertainty. Each patch is then classified as accurate and certain (AC), accurate and uncertain (AU), inaccurate and certain (IC), or inaccurate and uncertain (IU). PAvPU is computed as
    
    \begin{equation} \mathrm{PAvPU} = \frac{N_{AC} + N_{IU}}{N_{AC} + N_{AU} + N_{IC} + N_{IU}}, \end{equation} 
    
    where $N_{AC}$, $N_{AU}$, $N_{IC}$ and $N_{IU}$ are the corresponding patch counts. A patch is considered accurate when at least half of its points are correctly classified, and the median patch uncertainty of the test split is used as the uncertainty threshold. Since the uncertainty map is model independent, this threshold is the same for all model variants within a dataset. PAvPU values are therefore compared only within each dataset.

    Because the threshold is the median, exactly half of the patches are labelled uncertain for every model, so if $\alpha$ denotes the fraction of accurate patches, PAvPU is bounded above by $1 - |\alpha - 0.5|$. A highly accurate model therefore cannot reach a high PAvPU, since at least $\alpha - 0.5$ of all patches must fall into AU, and PAvPU should be read together with accuracy, particularly for the stronger models. Region-level alignment can also be measured by rank agreement, as in the Ranking IoU of~\cite{qi2023nsa}, which ranks voxels separately by mean prediction error and by mean uncertainty and compares the resulting sequences. We adopt PAvPU because its four categories separate the cases that matter differently in practice. A model that is inaccurate and certain is failing silently, whereas one that is inaccurate and uncertain is signalling its own unreliability, and a single rank correlation value cannot distinguish between them.

\end{itemize}

While accuracy provides an overall measure of correctness, mIoU gives greater weight to performance on minority classes~\cite{guo2023mctnet,everingham2010pascal}, and PAvPU measures the alignment between uncertainty and prediction errors. We therefore use mIoU and PAvPU as the primary metrics in this study.

\section{Datasets} \label{sec:datasets}
For the experiments, four benchmark LiDAR point cloud datasets have been utilised. These datasets provide outdoor large-scale, annotated point clouds captured from different sensing platforms, thereby covering a wide variety of urban environments and scene complexities.

\subsection{Vaihingen Dataset}
The Vaihingen 3D labeling benchmark dataset~\cite{rottensteiner2012isprs}, released by the International Society for Photogrammetry and Remote Sensing, contains ALS point clouds covering an urban area in Vaihingen, Germany. Two areas are utilised in this work, Area 2 with 266,675 points and Area 3 with 323,896 points, with Area 2 used for training and Area 3 held out for testing. Nine classes are defined for this dataset, but only four are present in the held-out Area 3, namely low vegetation, impervious surfaces, roof and tree, and mIoU is averaged over those four.

\subsection{AHN3 Dataset}
The Actueel Hoogtebestand Nederland~\cite{ahn3} is a high-resolution ALS dataset that provides detailed height measurements across the entire Netherlands. The data are captured using airborne LiDAR from helicopters and aircraft, achieving a vertical accuracy of approximately 5 cm, and the AHN3 release covers the period from 2014 to 2019. We use three tiles, with 28AN1\_16 and 28AZ1\_13 for training and 32CN2\_11 for testing, and a label set comprising never classified, vegetation, ground, building, water, and a reserved class covering bridges and similar structures. The test tile comprises approximately 17.2 million points.

\subsection{Paris-Lille-3D Dataset}
The Paris-Lille-3D dataset~\cite{roynard2018paris} is a large-scale MLS benchmark containing more than 140 million hand-labeled points collected from two cities in France. Two versions are available, a 50 class model and a 10 class model, and in this work the 10 class benchmark is used. It consists of nine annotated semantic categories, namely ground, building, pole or road sign or traffic light, bollard or small pole, trash can, barrier, pedestrian, car and natural vegetation, along with one unclassified class. We train on the three Lille sections and test on Paris, comprising approximately 38.3 million points, which constitutes a genuine cross-city transfer.

\subsection{Toronto-3D Dataset}
The Toronto-3D dataset~\cite{tan2020toronto} consists of 78.3 million points covering four urban road sections of approximately 250 m each, captured using MLS. Each point is enriched with multiple attributes, including coordinates, RGB color, intensity, GPS time, scan angle rank and semantic labels. The dataset defines eight annotated classes, namely road, road markings, natural objects such as trees and shrubs, buildings, utility lines, poles, cars and fences, along with one unclassified category. We train on L001, L002 and L004, and test on L003, comprising approximately 39.7 million points. The UTM offset is subtracted before feature computation, since retaining it produces numerically unstable neighborhood statistics.

\section{Experimental Setup} \label{sec:setup}
\textbf{System side:} The experimental setup consisted of a workstation powered by an AMD Ryzen 9 7950X 16-core processor (4.5 to 5.7 GHz), 128 GB RAM, and an NVIDIA GeForce RTX 4080 SUPER GPU with 16 GB GDDR6X memory. The environment was configured on Windows 11 Pro using WSL with Ubuntu 22.04 LTS.

\textbf{Model side:} The training configuration of all models is summarized in Table~\ref{tab:hyperparams}. All ANN runs are driven from a single configuration, so that no hyperparameter is specified separately per experiment. The optimizer is Adam, the batch size is 512, and training runs for 30 epochs. The random seed was selected from 42, 43, and 44 on the basis of test mIoU, and an optimal seed of 42 is used for all ANN experiments. The architecture uses hidden layers of size 128, 64, and 32 with ReLU activation and no dropout. The covariance regularization for the Gaussian model is $\epsilon = 10^{-3}$, coordinates are excluded from the uncertainty model, and $K = 30$ is used for PAvPU patches. No validation slice is carved out of the training data for any model, and the model after the final epoch is evaluated, since the canonical spatial test split is the only properly held-out set and no decision during training should depend on it. The choice of the ANN seed is the only exception. The three variants share identical optimization settings, so that any difference between them is attributable to the uncertainty treatment. PointNet++ and RandLA-Net are trained on spatial blocks drawn at random from the training files, 20 m blocks of 40,960 points for the two mobile datasets and 64 m blocks of 32,768 points for AHN3, and are evaluated on overlapping blocks so that every test point is predicted several times and the softmax outputs are summed. Both architectures use the same block sampler and are therefore trained on identical blocks, and each baseline shares its settings with its uncertainty-augmented counterpart, so that the comparison within each family remains attributable to the uncertainty treatment as well. For both reference models, no seed is selected. Every configuration is trained with each of the three seeds, and the reported values are the mean over the three runs.

\begin{table*}[t]
  \centering
  \caption{Training configuration of the three model families. Each baseline shares its configuration with its uncertainty-augmented counterpart.}
  \label{tab:hyperparams}
  \small
  \renewcommand{\arraystretch}{1.15}
  \begin{tabular}{llll}
    \toprule
    Parameter & ANN (M1, M2, M3) & PointNet++ & RandLA-Net \\
    \midrule
    Input channels   & 17 (M2: 18)   & 3 (+1 with $\tilde{u}_i$)  & 3 (+1 with $\tilde{u}_i$) \\
    Optimizer        & Adam          & Adam                       & Adam \\
    Learning rate    & $10^{-3}$     & $10^{-3}$, cosine schedule & $10^{-2}$, decay 0.95 per epoch \\
    Weight decay     & none          & $10^{-4}$                  & none \\
    Batch size       & 512 points    & 8 blocks (Vaihingen: 4)    & 4 blocks \\
    Epochs           & 30            & 50 (2000 blocks each, Vaihingen: 1000) & 50 (2000 blocks each, Vaihingen: 1000) \\
    Dropout          & none          & 0.5                        & 0.5 \\
    \bottomrule
  \end{tabular}
\end{table*}

\textbf{Data side:} For Paris-Lille-3D, with over 80 million training points, the training data is streamed in batches rather than materialized as framework tensors, and integer class indices with a sparse cross-entropy formulation are used in place of one-hot label matrices. Both are memory management choices with no effect on the objective.

\begin{table}[t]
  \centering
    \caption{Effect of the neighborhood radius on test accuracy and on the share of test points   without valid descriptors (dropped), per dataset. Each configuration was trained for five   epochs. The radius was selected on the training splits, where $r = 0.25$~m also drops the fewest points on every dataset; test values are reported for reference. The most optimal values are in boldface}
  \label{tab:radius_effect}
  \small
  \renewcommand{\arraystretch}{1.15}
  \begin{tabular}{llcc}
    \toprule
    Dataset & $r$ (m) & Test Acc. & Dropped Pts (\%) \\
    \midrule
    \multirow{4}{*}{Vaihingen}
      & 0.10 & 0.2304 & 0.13 \\
      & 0.15 & 0.2263 & 0.02 \\
      & 0.20 & 0.2804 & 0.00 \\
      & \textbf{0.25} & \textbf{0.3155} & \textbf{0.00} \\
    \midrule
    \multirow{4}{*}{AHN3}
      & 0.10 & 0.3699 & 93.47 \\
      & 0.15 & 0.3128 & 83.91 \\
      & 0.20 & 0.3418 & 72.31 \\
      & \textbf{0.25} & \textbf{0.3685} & \textbf{60.85} \\
    \midrule
    \multirow{4}{*}{Paris-Lille-3D}
      & 0.10 & 0.4424 & 0.88 \\
      & 0.15 & 0.4973 & 0.27 \\
      & 0.20 & 0.3321 & 0.11 \\
      & \textbf{0.25} & \textbf{0.6601} & \textbf{0.05} \\
    \midrule
    \multirow{4}{*}{Toronto-3D}
      & 0.10 & 0.7962 & 0.71 \\
      & 0.15 & 0.7984 & 0.29 \\
      & 0.20 & 0.8023 & 0.15 \\
      & \textbf{0.25} & \textbf{0.8224} & \textbf{0.08} \\
    \bottomrule
  \end{tabular}
\end{table}

To analyze the effect of local context, classification was performed at four different neighborhood radii, namely 0.10 m, 0.15 m, 0.20 m and 0.25 m. Table~\ref{tab:radius_effect} presents the test accuracy and the share of dropped points across these radii for each dataset. A radius of 0.25 m yielded the most promising results and was therefore adopted for all subsequent experiments.

\section{Results} \label{sec:results}

\begin{figure*}[h!]
    \centering
    \includegraphics[width=\textwidth]{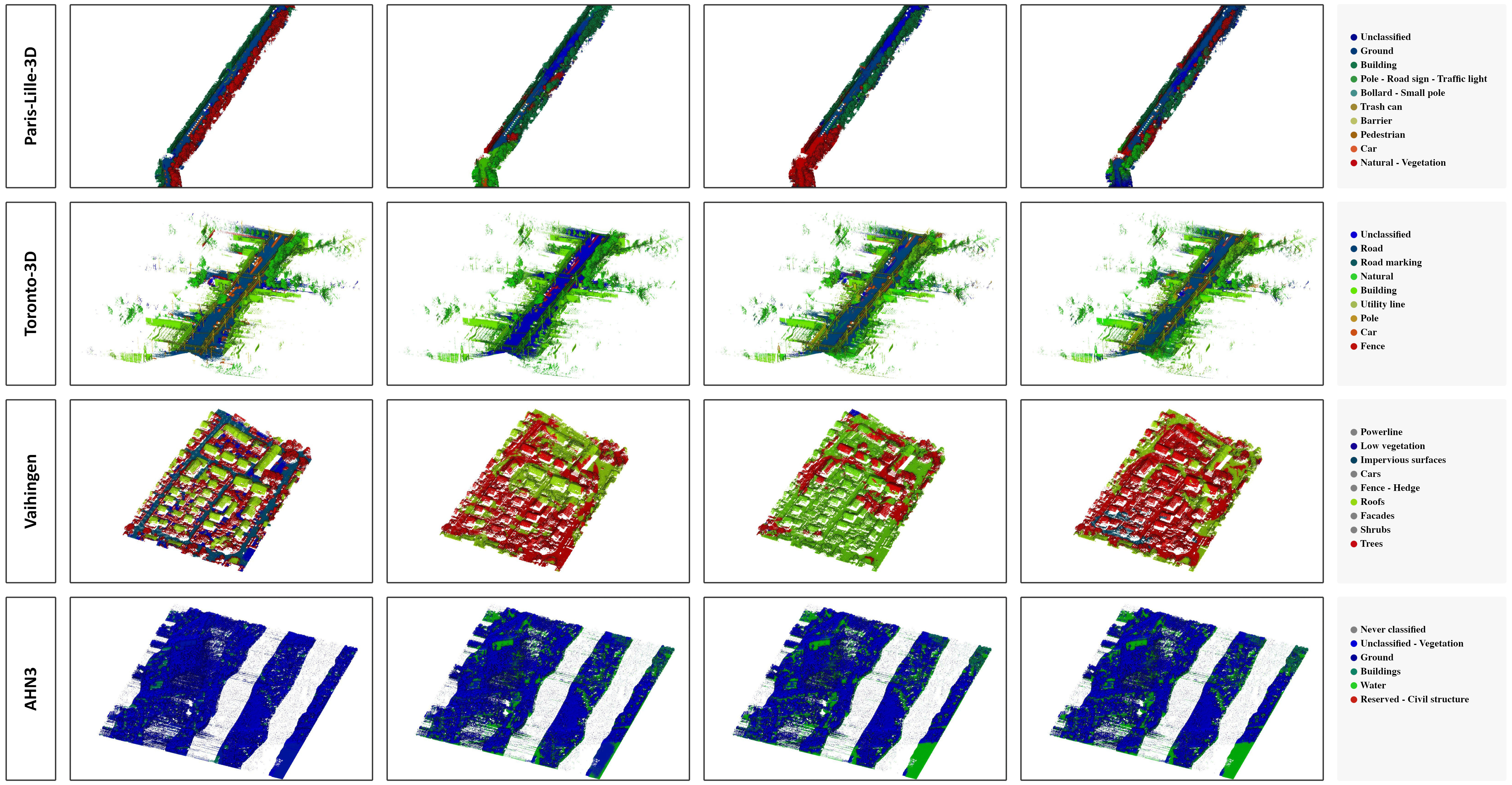}
    \caption{Segmentation ground truth (left) compared to predictions (M1, M2 and M3, respectively). Visualized using CloudCompare software.}
    \label{fig:gt_vs_pred}
\end{figure*}

\begin{figure*}[h!]
    \centering
    \includegraphics[width=\textwidth]{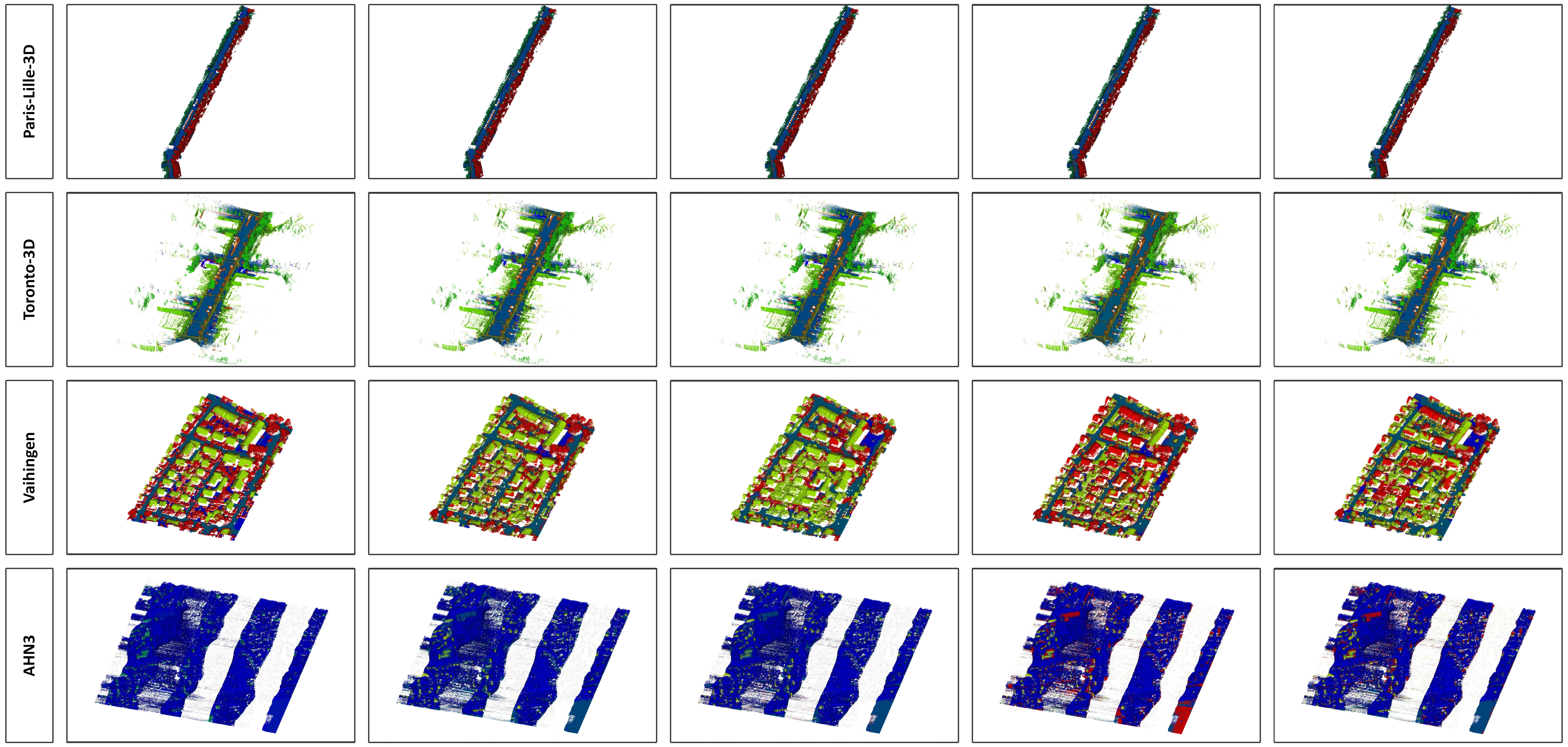}
    \caption{Segmentation ground truth (left) compared to the reference models (PointNet++ baseline and uncertainty-aware variant, RandLA-Net baseline and uncertainty-aware variant, respectively). For class details, refer to the legend in Fig.~\ref{fig:gt_vs_pred}. Visualized using CloudCompare software.}
    \label{fig:gt_vs_pred_pnpp}
\end{figure*}

\begin{table*}[t]
  \centering
  \caption{Test accuracy, mIoU and PAvPU across models and datasets under canonical
  spatially disjoint splits. Best value per dataset within each model family in boldface, second best within the ANN family underlined.}
  \label{tab:main}
  \renewcommand{\arraystretch}{1.25}
  \small
  \begin{tabular}{llccc}
    \toprule
    Dataset & Model & Test Acc. & mIoU & PAvPU \\

    \midrule
    {Vaihingen}
      & M1 Baseline ANN                  & \underline{0.3116} & \underline{0.1166} & \textbf{0.5066} \\
      & M2 ANN + uncertainty feature     & 0.2856 & 0.1095 & \underline{0.4891} \\
      & M3 ANN + uncertainty loss        & \textbf{0.3322} & \textbf{0.1216} & 0.4770 \\

      \rowcolor{gray!15} \cellcolor{white}
      & PointNet++                       & 0.5489 & 0.3450 & 0.5148 \\
      \rowcolor{gray!15} \cellcolor{white}
      & PointNet++ + uncertainty feature & \textbf{0.6423} & \textbf{0.4275} & \textbf{0.5462} \\

      \rowcolor{gray!30} \cellcolor{white}
      & RandLA-Net                       & \textbf{0.5302} & 0.3037 & 0.5241 \\
      \rowcolor{gray!30} \cellcolor{white}
      & RandLA-Net + uncertainty feature & 0.5134 & \textbf{0.3197} & \textbf{0.5637} \\

    \midrule
    {AHN3}
      & M1 Baseline ANN                  & \textbf{0.4215} & \textbf{0.1566} & 0.5893 \\
      & M2 ANN + uncertainty feature     & \underline{0.3838} & \underline{0.1418} & \textbf{0.6081} \\
      & M3 ANN + uncertainty loss        & 0.3805 & 0.1399 & \underline{0.6048} \\

      \rowcolor{gray!15} \cellcolor{white}
      & PointNet++                       & 0.7762 & 0.3609 & \textbf{0.5602} \\
      \rowcolor{gray!15} \cellcolor{white}
      & PointNet++ + uncertainty feature & \textbf{0.7928} & \textbf{0.3697} & 0.5467 \\

      \rowcolor{gray!30} \cellcolor{white}
      & RandLA-Net                       & 0.5875 & 0.2782 & \textbf{0.6292} \\
      \rowcolor{gray!30} \cellcolor{white}
      & RandLA-Net + uncertainty feature & \textbf{0.6103} & \textbf{0.2968} & 0.6219 \\

    \midrule
    {Paris-Lille-3D}
      & M1 Baseline ANN                  & 0.3611 & \underline{0.1009} & \underline{0.5438} \\
      & M2 ANN + uncertainty feature     & \underline{0.3883} & \textbf{0.1126} & 0.5242 \\
      & M3 ANN + uncertainty loss        & \textbf{0.4066} & 0.0966 & \textbf{0.7024} \\

      \rowcolor{gray!15} \cellcolor{white}
      & PointNet++                       & 0.9467 & 0.6662 & \textbf{0.5431} \\
      \rowcolor{gray!15} \cellcolor{white}
      & PointNet++ + uncertainty feature & \textbf{0.9507} & \textbf{0.6771} & 0.5392 \\

      \rowcolor{gray!30} \cellcolor{white}
      & RandLA-Net                       & \textbf{0.9430} & 0.5593 & 0.5438 \\
      \rowcolor{gray!30} \cellcolor{white}
      & RandLA-Net + uncertainty feature & 0.9429 & \textbf{0.5603} & \textbf{0.5468} \\

    \midrule
    {Toronto-3D}
      & M1 Baseline ANN                  & \textbf{0.8571} & 0.3457 & 0.5587 \\
      & M2 ANN + uncertainty feature     & \underline{0.8026} & \textbf{0.3745} & \underline{0.5933} \\
      & M3 ANN + uncertainty loss        & 0.7755 & \underline{0.3721} & \textbf{0.6054} \\

      \rowcolor{gray!15} \cellcolor{white}
      & PointNet++                       & 0.9318 & 0.7179 & \textbf{0.4896} \\
      \rowcolor{gray!15} \cellcolor{white}
      & PointNet++ + uncertainty feature & \textbf{0.9332} & \textbf{0.7257} & 0.4883 \\

      \rowcolor{gray!30} \cellcolor{white}
      & RandLA-Net                       & 0.9262 & 0.6451 & \textbf{0.4957} \\
      \rowcolor{gray!30} \cellcolor{white}
      & RandLA-Net + uncertainty feature & \textbf{0.9275} & \textbf{0.6506} & 0.4940 \\

    \bottomrule
  \end{tabular}
\end{table*}

\begin{table*}[t]
  \centering
  \caption{Per-class IoU on Toronto-3D (Tested on L003). The results clearly show that the uncertainty aware ANN variants are capable of recovering minority classes at a cost concentrated in the majority classes. Unclassified is excluded from mIoU throughout this table. Best value per class within each model family in boldface, second best within the ANN family underlined.}
  \label{tab:toronto_perclass}
  \small
  \resizebox{0.90\textwidth}{!}{%
  \begin{tabular}{lrccc>{\columncolor{gray!15}}c>{\columncolor{gray!15}}c>{\columncolor{gray!30}}c>{\columncolor{gray!30}}c}
    \toprule
    Class & \# test points & M1 & M2 & M3 & PointNet++ & PointNet++ + unc. & RandLA-Net & RandLA-Net + unc. \\
    \midrule
    road          & 20{,}557{,}596 & \textbf{0.8943} & \underline{0.8712} & 0.8612 & 0.9037 & \textbf{0.9055} & 0.9057 & 0.9057 \\
    building      & 11{,}676{,}403 & \textbf{0.7921} & \underline{0.7236} & 0.6876 & 0.9580 & \textbf{0.9610} & 0.9407 & \textbf{0.9438} \\
    car           &  2{,}370{,}844 & \textbf{0.6166} & \underline{0.4137} & 0.3630 & 0.9229 & \textbf{0.9243} & \textbf{0.9258} & 0.9187 \\
    natural       &  1{,}939{,}505 & \textbf{0.4547} & \underline{0.3848} & 0.3588 & 0.9118 & \textbf{0.9248} & 0.8533 & \textbf{0.8850} \\
    road marking  &    785{,}591 & 0.0000 & \underline{0.0060} & \textbf{0.0123} & 0.0001 & \textbf{0.0002} & 0.0000 & 0.0000 \\
    pole          &    407{,}843 & 0.0000 & \textbf{0.2803} & \underline{0.2613} & 0.8803 & \textbf{0.8873} & \textbf{0.8309} & 0.8054 \\
    utility line  &    337{,}735 & 0.0077 & \underline{0.3161} & \textbf{0.4289} & \textbf{0.8120} & 0.8067 & 0.7040 & \textbf{0.7408} \\
    fence         &    264{,}047 & 0.0000 & \underline{0.0001} & \textbf{0.0035} & 0.3542 & \textbf{0.3959} & 0.0001 & \textbf{0.0053} \\
    \midrule
    \textbf{mIoU} & & 0.3457 & \textbf{0.3745} & \underline{0.3721} & 0.7179 & \textbf{0.7257} & 0.6451 & \textbf{0.6506} \\
    \bottomrule
  \end{tabular}%
  }
\end{table*}

All experiments were conducted with features computed using the \texttt{Jakteristics} library at a search radius of 0.25 m, under the canonical spatially disjoint splits of Table~\ref{tab:TrainTestSplit}. A comparison of all seven models across the four datasets, reporting accuracy, mIoU and PAvPU, is given in Table~\ref{tab:main}, and Figures~\ref{fig:gt_vs_pred} and~\ref{fig:gt_vs_pred_pnpp}. The baseline models also serve as ablation studies for the proposed roles of determinstic epistemic uncertainty in the training models.

\subsection{Overall Trends}
We make the following three observations across all datasets.

\textbf{First}, test accuracy varies considerably across datasets, ranging from 0.29 to 0.95, and even for PointNet++, the strongest model in our study, it remains below 0.80 for both the ALS datasets. This is expected in a spatially disjoint evaluation. A model trained on one region transfers imperfectly to a new region, where the descriptor distribution can shift with scene composition, point density, and acquisition geometry\cite{jiang2022groundaware}.

\textbf{Second}, uncertainty-aware modeling produces little consistent change in the overall accuracy. Within the ANN family, the uncertainty-weighted loss gives the highest accuracy on Vaihingen and Paris-Lille-3D, while the baseline performs better on AHN3 and Toronto-3D. RandLA-Net shows the same inconsistency, with accuracy falling for Vaihingen, rising for AHN3 and Toronto-3D, and remaining unchanged for Paris-Lille-3D. PointNet++ is the exception, with accuracy rising for all four datasets, although the gain is substantial only for Vaihingen, \ie~an increase from 0.5489 to 0.6423. Since accuracy on these datasets is dominated by a small number of majority classes, we do not interpret the smaller changes as meaningful on their own.

\textbf{Third}, in contrast, mIoU and PAvPU show a more consistent improvement with uncertainty-aware modeling. mIoU improves over the baseline on Vaihingen, Paris-Lille-3D and Toronto-3D, while PAvPU improves for AHN3, Paris-Lille-3D and Toronto-3D. The improvement is often substantial. For Toronto-3D, mIoU rises from 0.3457 to 0.3745 while accuracy falls, and for Paris-Lille-3D, PAvPU rises from 0.5438 to 0.7024. Section~\ref{sec:perclass} examines the class-wise mIoU gains, while Section~\ref{sec:pavpuresults} examines the corresponding PAvPU results.

The same pattern appears for mIoU with the PointNet++ reference model. Supplying $\tilde{u}_i$ as an additional input channel improves mIoU on all four datasets. The gain is large on Vaihingen, where mIoU rises from 0.3450 to 0.4275, and small on Paris-Lille-3D and Toronto-3D, from 0.6662 to 0.6771 and from 0.7179 to 0.7257. On AHN3, the rise from 0.3609 to 0.3697 is not considered reliable, since ground segmentation on this dataset is unstable. PAvPU, in contrast, improves only on Vaihingen, from 0.5148 to 0.5462, and decreases slightly on the other three datasets.

RandLA-Net shows the same trend for mIoU. Supplying $\tilde{u}_i$ improves mIoU on all four datasets, with the largest gains on the two airborne datasets, where mIoU rises from 0.3037 to 0.3197 on Vaihingen and from 0.2782 to 0.2968 on AHN3. On Paris-Lille-3D and Toronto-3D, where the baseline already exceeds 0.92 accuracy, the gains are marginal. PAvPU is mixed, improving on Vaihingen from 0.5241 to 0.5637 and slightly on Paris-Lille-3D, while decreasing slightly on AHN3 and Toronto-3D. As discussed in Section~\ref{sec:pavpuresults}, PAvPU for both reference models on the two mobile datasets is limited by the median threshold rather than by the uncertainty estimate.

\emph{The improvement in mIoU across all three architectures, including the strongest one, suggests that incorporating uncertainty into the classification process has a positive effect that is not specific to the point-wise network or to models of limited capacity.}

\subsection{Class-wise Behavior} \label{sec:perclass}
The class-wise breakdown explains why mIoU and accuracy move in opposite directions. Table~\ref{tab:toronto_perclass} gives class-wise IoU on Toronto-3D, which is the clearest case.

The baseline ANN performs poorly on minority classes. It has largely collapsed onto the majority classes, a common effect of severe class imbalance that overall accuracy can hide~\cite{aditya2024vegetation}. Both uncertainty-aware variants reduce this collapse. Utility line IoU rises from 0.0077 to 0.3161 with M2 and 0.4289 with M3, while pole rises from 0 to 0.2803 and 0.2613, respectively. These thin and geometrically atypical structures are also the classes assigned higher uncertainty by Equation~\ref{eq:uncertainty}, suggesting that the uncertainty signal is particularly useful for recovering minority classes. The resulting accuracy drop comes from major classes, explaining why accuracy falls while mIoU improves.

For both reference models, the baseline already recovers most minority classes, with pole at 0.8803 and 0.8309 and utility line at 0.8120 and 0.7040 for PointNet++ and RandLA-Net, respectively, so the uncertainty feature has less room to act. For RandLA-Net, it still improves utility line to 0.7408 and natural to 0.8850, while pole and car decrease slightly, to 0.8054 and 0.9187. For PointNet++, it gives small gains on every class except utility line, the largest for fence, which rises from 0.3542 to 0.3959 and is not recovered by any other model. Road marking is not recovered by any reference model and only marginally by the uncertainty-aware ANN variants. Road markings are geometrically identical to the surrounding road and are distinguished mainly by intensity, which is discarded during data preparation, so no geometric uncertainty signal can separate them.

The same pattern, in weaker form, holds on Paris-Lille-3D, as shown in Table~\ref{tab:paris_perclass}. Barrier improves from 0.0114 to 0.0513 and natural vegetation from 0.0967 to 0.1624 under M2. Ground improves from 0.3722 to 0.4719 under M3.

For RandLA-Net, the class-wise changes on Paris-Lille-3D are mixed and largely cancel in mIoU. Bollard improves from 0.2642 to 0.3521 and car from 0.8254 to 0.8727, while pole, sign and light decreases from 0.6114 to 0.5322. For PointNet++, the changes are similarly mixed but small, with barrier improving from 0.5478 to 0.5943 and pedestrian from 0.4535 to 0.4814, while car and bollard decrease slightly.

\begin{table*}[t]
  \centering
  \caption{Per-class IoU on Paris-Lille-3D (Tested on Paris tile). Unclassified   is excluded from mIoU throughout this table. Best value per class within each model family in boldface, second best within the ANN family underlined.}
  \label{tab:paris_perclass}
  \small
  \resizebox{0.90\textwidth}{!}{%
  \begin{tabular}{lrccc>{\columncolor{gray!15}}c>{\columncolor{gray!15}}c>{\columncolor{gray!30}}c>{\columncolor{gray!30}}c}
    \toprule
    Class & \# test points & M1 & M2 & M3 & PointNet++ & PointNet++ + unc. & RandLA-Net & RandLA-Net + unc. \\
    \midrule
    ground              & 18{,}103{,}476 & 0.3722 & \underline{0.3828} & \textbf{0.4719} & \textbf{0.9872} & 0.9871 & \textbf{0.9901} & 0.9893 \\
    building            &  8{,}946{,}816 & \textbf{0.4001} & \underline{0.3677} & 0.2207 & 0.9173 & \textbf{0.9296} & \textbf{0.9171} & 0.9071 \\
    natural vegetation  &  6{,}198{,}638 & 0.0967 & \textbf{0.1624} & \underline{0.1500} & 0.8852 & \textbf{0.8960} & \textbf{0.8777} & 0.8669 \\
    barrier             &  2{,}389{,}847 & \underline{0.0114} & \textbf{0.0513} & 0.0013 & 0.5478 & \textbf{0.5943} & \textbf{0.4360} & 0.4275 \\
    car                 &  1{,}298{,}478 & 0.0111 & \textbf{0.0283} & \underline{0.0148} & \textbf{0.8770} & 0.8668 & 0.8254 & \textbf{0.8727} \\
    pole / sign / light &    193{,}166 & \textbf{0.0155} & \underline{0.0143} & 0.0109 & 0.6666 & \textbf{0.6730} & \textbf{0.6114} & 0.5322 \\
    pedestrian          &    137{,}272 & 0.0000 & 0.0000 & 0.0000 & 0.4535 & \textbf{0.4814} & 0.0000 & 0.0000 \\
    trash can           &     28{,}113 & 0.0000 & 0.0000 & 0.0000 & 0.1851 & \textbf{0.2001} & \textbf{0.1114} & 0.0949 \\
    bollard / small pole &     24{,}327 & \underline{0.0010} & \textbf{0.0064} & 0.0002 & \textbf{0.4759} & 0.4654 & 0.2642 & \textbf{0.3521} \\
    \midrule
    \textbf{mIoU} & & \underline{0.1009} & \textbf{0.1126} & 0.0966 & 0.6662 & \textbf{0.6771} & 0.5593 & \textbf{0.5603} \\
    \bottomrule
  \end{tabular}%
  }
\end{table*}

\begin{table*}[t]
  \centering
  \caption{Per-class IoU on Vaihingen (trained on Area 2, tested on Area 3). Best value per class within each model family in boldface, second best within the ANN family underlined.}
  \label{tab:vaihingen_perclass}
  \small
  \resizebox{0.90\textwidth}{!}{%
  \begin{tabular}{lrccc>{\columncolor{gray!15}}c>{\columncolor{gray!15}}c>{\columncolor{gray!30}}c>{\columncolor{gray!30}}c}
    \toprule
    
    Class & \# test points & M1 & M2 & M3 & PointNet++ & PointNet++ + unc. & RandLA-Net & RandLA-Net + unc. \\
    \midrule
    
    Tree                 & 135{,}449 & \underline{0.2748} & 0.2430 & \textbf{0.3250} & 0.3460 & \textbf{0.4517} & \textbf{0.4155} & 0.3753 \\
    
    Impervious surfaces  &  83{,}372 & 0.0000 & \underline{0.0148} & \textbf{0.0235} & 0.4467 & \textbf{0.5412} & \textbf{0.4818} & 0.4220 \\
    
    Roof                 &  69{,}563 & \textbf{0.1917} & \underline{0.1803} & 0.1380 & 0.4626 & \textbf{0.5559} & 0.2318 & \textbf{0.3132} \\
    
    Low vegetation       &  35{,}512 & 0.0000 & 0.0000 & \textbf{0.0001} & 0.1246 & \textbf{0.1613} & 0.0855 & \textbf{0.1682} \\
    \midrule
    
    \textbf{mIoU} & & \underline{0.1166} & 0.1095 & \textbf{0.1216} & 0.3450 & \textbf{0.4275} & 0.3037 & \textbf{0.3197} \\
    \bottomrule
  
  \end{tabular}%
  }
\end{table*}

\begin{table*}[t]
  \centering
  \caption{Per-class IoU on AHN3 (trained on tiles 28AN1\_16 and 28AZ1\_13,   tested on 32CN2\_11). The never-classified category has no points in the test   tile and is omitted. Best value per class within each model family in boldface, second best within the ANN family underlined.}
  \label{tab:ahn3_perclass}
  \footnotesize
  \resizebox{0.90\textwidth}{!}{%
  \begin{tabular}{lrccc>{\columncolor{gray!15}}c>{\columncolor{gray!15}}c>{\columncolor{gray!30}}c>{\columncolor{gray!30}}c}
    \toprule
    Class & \# test points & M1 & M2 & M3 & PointNet++ & PointNet++ + unc. & RandLA-Net & RandLA-Net + unc. \\
    \midrule
    
    Vegetation & 9{,}377{,}068 & 0.6720 & \textbf{0.6884} & \underline{0.6824} & \textbf{0.9257} & 0.9073 & 0.9493 & \textbf{0.9524} \\
    
    Ground     & 7{,}018{,}580 & \textbf{0.0984} & \underline{0.0002} & \underline{0.0002} & 0.5467 & \textbf{0.5795} & 0.0874 & \textbf{0.1346} \\
    
    Building   &   791{,}952 & 0.0126 & \textbf{0.0206} & \underline{0.0171} & 0.3323 & \textbf{0.3619} & 0.3543 & \textbf{0.3971} \\
    
    Reserved   &        38 & 0.0000 & 0.0000 & 0.0000 & 0.0000 & 0.0000 & 0.0000 & 0.0000 \\
    
    Water      &        23 & 0.0000 & 0.0000 & 0.0000 & 0.0000 & 0.0000 & 0.0000 & 0.0000 \\
    \midrule
    \textbf{mIoU} & & \textbf{0.1566} & \underline{0.1418} & 0.1399 & 0.3609 & \textbf{0.3697} & 0.2782 & \textbf{0.2968} \\
    \bottomrule
  \end{tabular}%
  }
\end{table*}

\begin{table*}[t]
  \centering
  \caption{PAvPU decomposition on the held out split. AC = accurate \& certain,   AU = accurate \& uncertain, IC = inaccurate \& certain, IU = inaccurate \&   uncertain, as percentages of all patches. AC $+$ AU is the share of accurate patches and is therefore not equal to the point-wise test accuracy. PAvPU $=$ AC $+$ IU. The   uncertainty threshold is the median patch uncertainty of the test split and is identical across all models. Best value per dataset within each model family in boldface, second best within the ANN family underlined. Higher is better for AC, IU and PAvPU, and lower is better for AU and IC.}
  \label{tab:pavpu_breakdown}
  \small
  \resizebox{0.74\textwidth}{!}{%
  \begin{tabular}{llrrrrc}
    \toprule
    Dataset & Model & AC (\%) & AU (\%) & IC (\%) & IU (\%) & PAvPU \\
    \midrule
    {Vaihingen}
      & M1 & \textbf{15.61} & \underline{14.95} & \textbf{34.39} & \underline{35.05} & \textbf{0.5066} \\
      & M2 & 13.71 & \textbf{14.80} & 36.29 & \textbf{35.20} & \underline{0.4891} \\
      & M3 & \underline{15.10} & 17.40 & \underline{34.90} & 32.60 & 0.4770 \\
      \rowcolor{gray!15} \cellcolor{white}
      & PointNet++ & 28.62 & \textbf{27.14} & 21.38 & \textbf{22.86} & 0.5148 \\
      \rowcolor{gray!15} \cellcolor{white}
      & PointNet++ + uncertainty & \textbf{35.41} & 30.79 & \textbf{14.59} & 19.21 & \textbf{0.5462} \\
      \rowcolor{gray!30} \cellcolor{white}
      & RandLA-Net & 28.20 & 25.79 & 21.80 & 24.21 & 0.5241 \\
      \rowcolor{gray!30} \cellcolor{white}
      & RandLA-Net + uncertainty & \textbf{29.37} & \textbf{23.01} & \textbf{20.63} & \textbf{26.99} & \textbf{0.5637} \\
    \midrule
    {AHN3}
      & M1 & \textbf{25.57} & 16.65 & \textbf{24.43} & 33.35 & 0.5893 \\
      & M2 & \underline{24.61} & \underline{13.80} & \underline{25.39} & \underline{36.20} & \textbf{0.6081} \\
      & M3 & 24.19 & \textbf{13.71} & 25.81 & \textbf{36.29} & \underline{0.6048} \\
      \rowcolor{gray!15} \cellcolor{white}
      & PointNet++ & 42.45 & \textbf{36.44} & 7.55 & \textbf{13.56} & \textbf{0.5602} \\
      \rowcolor{gray!15} \cellcolor{white}
      & PointNet++ + uncertainty & \textbf{42.53} & 37.86 & \textbf{7.47} & 12.14 & 0.5467 \\
      \rowcolor{gray!30} \cellcolor{white}
      & RandLA-Net & 35.39 & \textbf{22.47} & 14.61 & \textbf{27.53} & \textbf{0.6292} \\
      \rowcolor{gray!30} \cellcolor{white}
      & RandLA-Net + uncertainty & \textbf{36.11} & 23.92 & \textbf{13.89} & 26.08 & 0.6219 \\
    \midrule
    {Paris-Lille-3D}
      & M1 & 20.31 & \underline{15.93} & 29.69 & \underline{34.07} & \underline{0.5438} \\
      & M2 & \underline{20.66} & 18.24 & \underline{29.34} & 31.76 & 0.5242 \\
      & M3 & \textbf{30.54} & \textbf{10.30} & \textbf{19.46} & \textbf{39.70} & \textbf{0.7024} \\
      \rowcolor{gray!15} \cellcolor{white}
      & PointNet++ & 49.74 & \textbf{45.42} & 0.27 & \textbf{4.58} & \textbf{0.5431} \\
      \rowcolor{gray!15} \cellcolor{white}
      & PointNet++ + uncertainty & \textbf{49.75} & 45.83 & \textbf{0.25} & 4.17 & 0.5392 \\
      \rowcolor{gray!30} \cellcolor{white}
      & RandLA-Net & 49.51 & 45.13 & 0.49 & 4.87 & 0.5438 \\
      \rowcolor{gray!30} \cellcolor{white}
      & RandLA-Net + uncertainty & \textbf{49.70} & \textbf{45.02} & \textbf{0.30} & \textbf{4.98} & \textbf{0.5468} \\
    \midrule
    {Toronto-3D}
      & M1 & \textbf{45.99} & 40.12 & \textbf{4.01} & 9.88 & 0.5587 \\
      & M2 & \underline{44.94} & \underline{35.61} & \underline{5.06} & \underline{14.39} & \underline{0.5933} \\
      & M3 & 44.18 & \textbf{33.64} & 5.82 & \textbf{16.36} & \textbf{0.6054} \\
      \rowcolor{gray!15} \cellcolor{white}
      & PointNet++ & 46.14 & \textbf{47.18} & 3.86 & \textbf{2.82} & \textbf{0.4896} \\
      \rowcolor{gray!15} \cellcolor{white}
      & PointNet++ + uncertainty & 46.14 & 47.31 & 3.86 & 2.69 & 0.4883 \\
      \rowcolor{gray!30} \cellcolor{white}
      & RandLA-Net & \textbf{46.17} & \textbf{46.60} & \textbf{3.83} & \textbf{3.40} & \textbf{0.4957} \\
      \rowcolor{gray!30} \cellcolor{white}
      & RandLA-Net + uncertainty & 46.16 & 46.76 & 3.84 & 3.24 & 0.4940 \\
    \bottomrule
  \end{tabular}%
  }
\end{table*}

\begin{figure*}[h!]
    \centering
    \includegraphics[width=\textwidth]{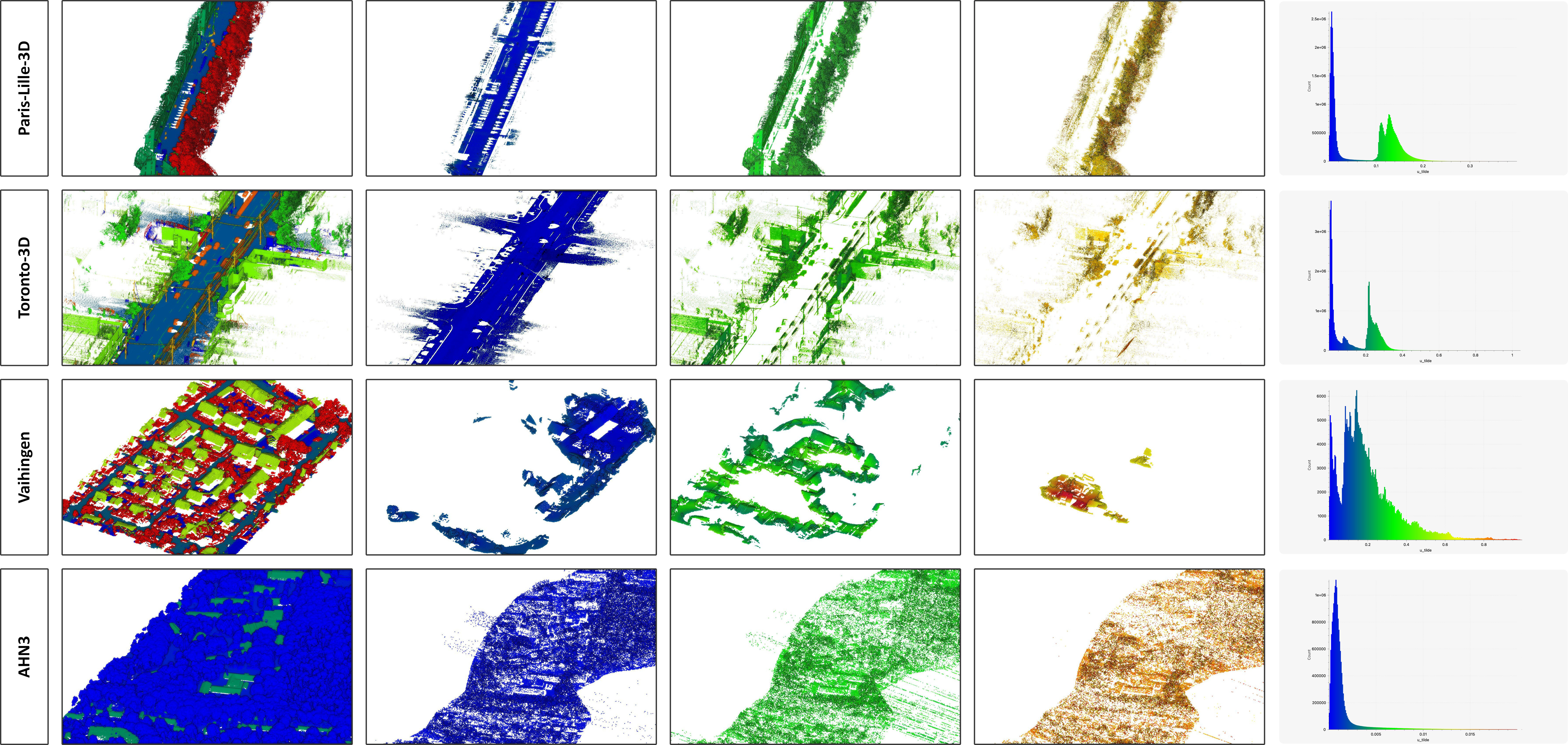}
    \caption{Uncertainty $\tilde{u}$ on the test tiles. Column~1 shows the ground truth labels. Columns~2 to~4 show subsets of points in the low (blue), intermediate (green) and high (yellow to red) ranges of $\tilde{u}$, and column~5 shows the histogram of $\tilde{u}$ with the same color scale, from which the value range of each subset can be read. The histogram tails were clipped per dataset before mapping the color scale, so the ranges follow each dataset's distribution, and hence, the scales differ between rows. High uncertainty occurs in regions of vegetation, irregular geometry and class boundaries. Visualized using CloudCompare.}
    \label{fig:findings}
\end{figure*}

Paris-Lille-3D also exposes the limits of the point-wise ANN approach. The pedestrian and trash can classes are never recovered by any ANN variant. Pedestrian is recovered only by PointNet++, with IoU of 0.4535 and 0.4814, and trash can by both reference models, with IoU of 0.1851 and 0.2001 for PointNet++ and 0.1114 and 0.0949 for RandLA-Net. These are compact and volumetrically ambiguous objects whose eigenvalue signature at 0.25 m is close to that of vegetation clutter, and no reweighting of a point-wise classifier over purely local descriptors can separate them without larger scale context. That they are recovered only by the reference models, which aggregate features over a progressively increasing receptive field, is consistent with this.

For the two airborne datasets, shown in Tables~\ref{tab:vaihingen_perclass} and~\ref{tab:ahn3_perclass}, RandLA-Net benefits from the uncertainty feature in a similar way. On Vaihingen, low vegetation, the smallest class in the test area, rises from 0.0855 to 0.1682 and roof from 0.2318 to 0.3132, while tree and impervious surfaces decrease, which is the same trade between minority and majority classes seen for the ANN on Toronto-3D. On AHN3, ground rises from 0.0874 to 0.1346 and building from 0.3543 to 0.3971, with vegetation essentially unchanged. PointNet++ benefits more clearly on Vaihingen, where the uncertainty feature improves all four classes, with tree rising from 0.3460 to 0.4517, impervious surfaces from 0.4467 to 0.5412 and roof from 0.4626 to 0.5559, so that no trade between classes is observed. On AHN3, ground rises from 0.5467 to 0.5795 and building from 0.3323 to 0.3619, while vegetation decreases from 0.9257 to 0.9073, but ground segmentation on this dataset is unstable, so these changes are not considered reliable.

\subsection{Alignment Between Error and Uncertainty} \label{sec:pavpuresults}
Table~\ref{tab:pavpu_breakdown} breaks down PAvPU into its four patch categories and shows where the gains come from.

Within the ANN family, PAvPU improves with uncertainty-aware modeling on AHN3, Paris-Lille-3D and Toronto-3D, with the largest gain on Paris-Lille-3D, where M3 reaches 0.7024 compared with 0.5438 for the baseline. On Paris-Lille-3D, M3 reduces the inaccurate and certain patches from 29.69\% to 19.46\%, while increasing accurate and certain patches from 20.31\% to 30.54\%. A similar, though smaller, shift occurs on Toronto-3D, where the accurate and uncertain fraction falls by 6.5 points into the inaccurate and uncertain quadrant. This means more errors occur in regions that the uncertainty map already identifies as uncertain.

Since the uncertainty map is the same for all model variants, these changes come from where the models make their errors. Uncertainty-aware modeling shifts more errors towards regions that are already marked as geometrically atypical.

Vaihingen is the exception within the ANN family, with PAvPU decreasing from 0.5066 to 0.4891 and 0.4770. It is also by far the smallest dataset, with 0.32 million test points compared with 17 to 40 million for the others, and has the largest domain gap between the training and test areas. The point-based models behave differently on the same dataset. PAvPU improves from 0.5148 to 0.5462 for PointNet++, the only dataset on which its PAvPU improves, coming from a fall in inaccurate and certain patches from 21.38\% to 14.59\%. For RandLA-Net, PAvPU improves from 0.5241 to 0.5637, the largest PAvPU gain of RandLA-Net on any dataset, coming from a rise in inaccurate and uncertain patches from 24.21\% to 26.99\% and a fall in inaccurate and certain patches from 21.80\% to 20.63\%. Since the uncertainty map is identical for all models, the decrease for the ANN cannot be attributed to the uncertainty estimate alone.

\emph{These results suggest that, when training data is scarce, the benefit of the uncertainty estimate for error and uncertainty alignment depends on the capacity of the model to exploit it, rather than on the estimate itself.}

On Paris-Lille-3D and Toronto-3D, both reference models reach PAvPU values close to 0.5 despite accuracies above 0.92. This follows from the median threshold described in Section~\ref{sec:evalprotocol}. For RandLA-Net, with 94.6\% and 92.8\% of patches accurate, at least 44.6\% and 42.8\% of all patches must fall into the accurate and uncertain category, which bounds PAvPU at approximately 0.55 and 0.57. The same bound applies to PointNet++, and since the uncertainty feature raises its share of accurate patches further, the bound tightens, which is consistent with the slight decrease in its PAvPU on both datasets. Within this bound, the inaccurate and certain fraction is the more informative quantity. On Paris-Lille-3D it is only 0.49\% for the RandLA-Net baseline and 0.30\% with the uncertainty feature, and 0.27\% and 0.25\% for PointNet++, so that more than nine out of ten inaccurate patches are already flagged as uncertain.

\subsection{Uncertainty Visualization} \label{sec:visualization}
Across all evaluated datasets, uncertainty as a feature exhibits a strong correlation with semantic class characteristics and local geometric structure, as seen in Figure~\ref{fig:findings}. Classes with weak inherent geometry, high intra-class variability or sparse representation, such as vegetation, vehicles, pedestrians, thin poles and wires, consistently show higher epistemic uncertainty, and so do the boundaries between semantic regions. In contrast, geometrically stable and spatially coherent classes such as buildings, roads and ground surfaces demonstrate significantly lower uncertainty.

This behavior is most evident in dense datasets like Paris-Lille-3D and Toronto-3D. However, performance drops on the Vaihingen and AHN3 datasets. The low point density and small footprint of Vaihingen limit the model's reliability, while the point density variation within and between AHN3 tiles introduces significant noise into the uncertainty estimation, as discussed in Section~\ref{sec:ahn3}.

\section{Discussion} \label{sec:discussion}
In this section, we take a deep dive into a few critical observations from our experiments.

\begin{figure*}[t]
  \centering
  \includegraphics[width=\textwidth]{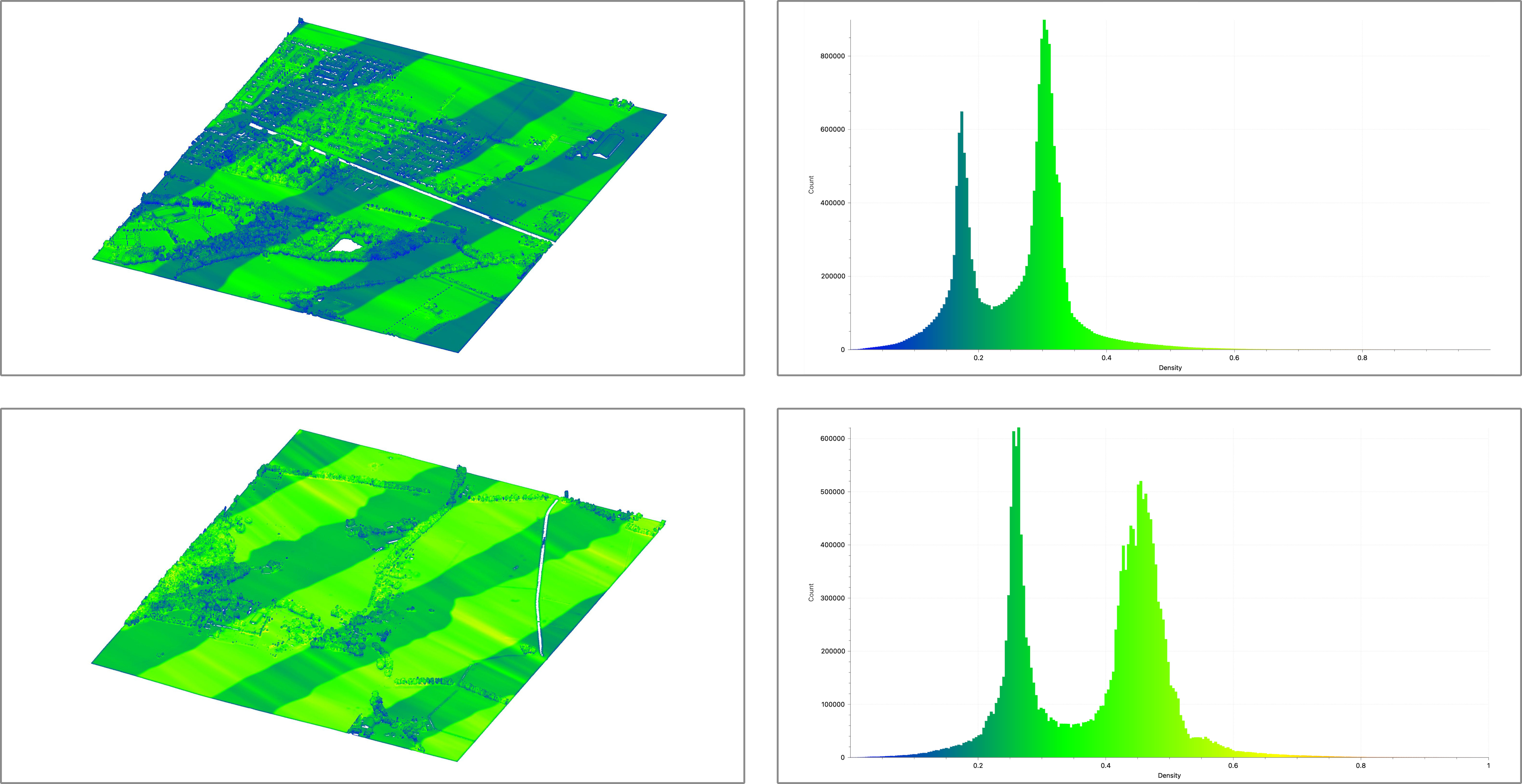}
  \caption{Point density variation within and between AHN3 tiles. Top: 28AN1\_16, Bottom: 28AZ1\_13. Density is the number of neighboring points within a 5\,m search radius, min-max normalised per tile to [0, 1]. Left: point clouds colored by normalised density, from blue (low) through green to yellow (high). Right: histograms of the normalised density, colored with the same scale. Within each tile, the bimodal distribution reflects the alternating bands of single flight-strip coverage (lower peak) and strip overlap (higher peak). Between tiles, the peaks lie at different values, so the same color does not represent the same absolute density in both tiles.}
  \label{fig:ahn3density}
\end{figure*}

\subsection{Why does AHN3 Underperform?} \label{sec:ahn3}
AHN3 is a unique dataset amongst the selected ones, on which uncertainty-aware modeling does not improve mIoU for the ANN family, and on which absolute performance is weak across all models. The per-class breakdown identifies two compounding causes, neither of which is a modeling failure.

\mysubheading{Class distribution shift between tiles} The held out tile 32CN2\_11 contains essentially no reserved class structures, with 38 points, whereas the training tiles contain them in quantity, and its class composition differs considerably from that of the training tiles. All three ANN variants nevertheless assign a very large fraction of the test tile to a class that is essentially absent from it, amounting to 7.77 million points for the baseline and 8.85 million for M2, while ground IoU collapses from 0.0984 under M1 to 0.0002 under M2 and M3. The models are confidently mislabeling ground. This is a shift in the label prior between acquisition tiles rather than a deficiency of the uncertainty estimate, because flat bare ground has nearly the same eigenvalue signature at a 0.25 m radius as other flat surfaces in the training tiles, and only contextual or intensity information distinguishes them. Ground IoU remains low for RandLA-Net as well, at 0.0874 for the baseline and 0.1346 with the uncertainty feature, despite its learned local representation and larger receptive field. Given the vegetation and building IoU, at most about two million ground points can be absorbed by those two classes, so more than four million ground points are assigned to classes that are essentially absent from the test tile, which is consistent with a shift in the label prior rather than a deficiency of the model. PointNet++ is the exception, with ground IoU of 0.5467 and 0.5795, which shows that this shift can be partly overcome by a learned spatial representation, although ground segmentation by PointNet++ on this dataset is unstable.

\mysubheading{Point density variation within tiles} AHN3 tiles are assembled from overlapping airborne flight lines, resulting in non uniform point density within a tile. Figure~\ref{fig:ahn3density} shows this clearly. The diagonal bands in both training tiles correspond to overlap regions between adjacent scan swaths, where surfaces are sampled by two passes rather than one. The resulting density is substantially higher in these regions than in the single pass regions between them. The histograms confirm this, with two distinct peaks in each tile corresponding to single pass and overlap coverage. Since this variation comes from the acquisition geometry rather than the semantic content, the same class can be observed at different point densities within a single tile.

This affects the geometric descriptors as well. Within a 0.25 m neighborhood, a change in point density changes both the number and spatial arrangement of neighboring points. The estimated local geometry can therefore differ substantially between overlap and single pass regions, even for the same physical structure. Each class may consequently occupy several regions of feature space rather than a single compact one. Fitting a single Gaussian per class, as in Equation~\ref{eq:gaussian}, absorbs this variation into a broader covariance, which makes the class conditional densities less distinct and weakens the uncertainty estimate. The same density variation also introduces changes in the classifier inputs that are unrelated to semantic content, which likely contributes to the low accuracy and mIoU observed on this dataset.

The very low median patch uncertainty on AHN3, at 0.0015 compared with 0.11 to 0.16 on the other datasets, is consistent with this explanation. With broad and overlapping class densities, many points receive similar log likelihood values, while a small number of outliers determine the training extrema. The normalized uncertainty is therefore compressed towards zero with a thin tail. Density variation between tiles, visible in Figure~\ref{fig:ahn3density} as a shift in the histogram peaks between the two training tiles, adds to this effect, but the variation within a tile is likely the larger factor.

Despite this compressed estimate, RandLA-Net still gains mIoU from the uncertainty channel on AHN3, improving ground from 0.0874 to 0.1346 and building from 0.3543 to 0.3971. For PointNet++, the mIoU gain on AHN3 is small and not reliable. The estimate therefore appears to remain informative on AHN3 when it is combined with a learned spatial representation, although less consistently than on the other datasets.

\subsection{Minority Class Recovery and the Accuracy Trade-off} \label{sec:tradeoff}
The per-class results in Tables~\ref{tab:toronto_perclass} and~\ref{tab:paris_perclass} show a consistent pattern. Uncertainty-aware modeling improves the classes that are sparsely represented and geometrically irregular, and the improvement is paid for by the majority classes. On Toronto-3D, utility line and pole are recovered from near zero while road, building and car decline, car most strongly, and since the majority classes account for most of the test points, overall accuracy falls even as mIoU improves. For the reference models the pattern is weaker on these two datasets, since their baselines already recover most minority classes. The clearest case for RandLA-Net is Vaihingen, where low vegetation, the smallest class, nearly doubles from 0.0855 to 0.1682 while tree and impervious surfaces decline. For PointNet++, no such trade-off is observed, as the uncertainty feature improves the minority classes largely without a cost in the majority classes, most clearly on Vaihingen, where all four classes improve.

This trade-off is also reported for an uncertainty-guided loss on Paris-Lille-3D in~\cite{qi2023nsa} as well, where the ground and building classes decline while minor classes improve substantially, across several backbones. That study isolates the effect by training a model to convergence without uncertainty and then enabling uncertainty-guided learning for further epochs. The explanation offered there is one of model capacity, in that resolving the boundaries of the dominant classes consumes capacity that becomes available to smaller objects once the contribution of uncertain points is reduced. Our results are consistent with this account, and add that the classes recovered are precisely those to which Equation~\ref{eq:uncertainty} assigns high uncertainty, since thin and irregular geometry is by construction atypical relative to the fitted class distributions.

The same mechanism explains why the two uncertainty-aware variants diverge on Paris-Lille-3D. Supplying uncertainty as a feature improves the minor classes, whereas weighting the loss by confidence reduces them below the baseline and lowers mIoU. Down-weighting by uncertainty suppresses exactly the points that carry the minority classes, because those classes are the high-uncertainty ones. M2 supplies the same signal as information rather than as suppression and does not have this failure mode. Both point-based models use the uncertainty in the same form as M2, and both improve mIoU on every dataset, although the gain for PointNet++ on AHN3 is not reliable, which is consistent with this reading. This also agrees with the argument in~\cite{yao2024uncertainty} that discarding low-reliability predictions wastes the information they carry, and suggests that treating high-uncertainty points as evidence to be used differently, rather than as evidence to be discounted, is the more promising direction. A weight that is bounded away from zero, as in~\cite{qi2023nsa}, would be a straightforward intermediate step.

\subsection{Limitations} \label{sec:limitations}
\begin{itemize}
    \item \textbf{Fixed neighborhood radius.} A radius of 0.25 m was used across all datasets and produced good results, despite the differences in point density and acquisition type. However, a fixed radius does not account for local variations in point density. As seen in Section~\ref{sec:ahn3}, an adaptive neighborhood size could better account for such variation and provide more consistent local geometric descriptors.
    
    \item \textbf{Point-wise model without broader spatial context.} The ANN classifier makes independent predictions for each point based on its local geometric descriptors. This limits its ability to use object level context, which can be important for classes such as pedestrians and trash cans. Both reference models, which aggregate context over a larger receptive field, recover trash can on Paris-Lille-3D, and PointNet++ also recovers pedestrians, which no ANN variant does.

    \item \textbf{Epistemic uncertainty only.} No aleatoric uncertainty is modeled.
    
    \item \textbf{PAvPU under a median threshold.} Since the uncertainty threshold is the median patch uncertainty, PAvPU is bounded above by $1 - |\alpha - 0.5|$, where $\alpha$ is the fraction of accurate patches. For highly accurate models such as both reference models on Paris-Lille-3D and Toronto-3D, this bound is close to 0.55, so PAvPU separates them poorly, and the inaccurate and certain fraction is the more informative quantity.

\end{itemize}

\subsection{Findings and Future Directions}
This study investigated the integration of uncertainty estimation into point cloud segmentation pipelines. The results highlight several findings.

\begin{itemize}

    \item Uncertainty-aware modeling improves mIoU and recovers minority classes that the baseline ANN never predicts at all, such as pole and utility line on Toronto-3D, at a cost concentrated in the majority classes. 

    \item It improves the alignment between prediction error and estimated uncertainty in half of the configurations. For the point-based models, the proportion of patches on which the model is confidently wrong falls in six of the eight cases, even where PAvPU itself is limited by the median threshold.
    
    \item The same uncertainty channel benefits PointNet++ and RandLA-Net as well as the point-wise network, improving mIoU on all four datasets for both architectures, which indicates that the signal is not specific to a single architecture or to models of limited capacity.

    \item On the smallest dataset, Vaihingen, the uncertainty estimate improves PAvPU for both point-based models but not for the ANN variants, which suggests that for scarce training data, the benefit of using uncertainty in the learning models depends on the capacity of the model to exploit the estimate rather than on the estimate itself.

    \item Uncertainty is consistently higher for classes with irregular geometry and for classes that are underrepresented in the data, and lower for geometrically stable and spatially coherent classes, which makes the estimate useful as a diagnostic overlay independently of any accuracy gain.
    
\end{itemize}

Several directions follow from these findings. Density adaptive or multiscale neighborhood selection could better handle the variation in point density observed in AHN3, and remove the need to use a single fixed neighborhood size across datasets. Estimating the Gaussians of Equation~\ref{eq:gaussian} in a network's learned latent space, rather than in the handcrafted descriptor space, could make the uncertainty estimate more closely tied to the model, although this would lose the model independence of the current approach. Adding an aleatoric component could help capture errors, such as the mislabeling of ground in AHN3, that are not explained by geometric atypicality alone. More spatially aware representations, including projection- and voxel-based approaches, could also provide broader spatial context that is difficult to capture with point-wise representations alone. An uncertainty threshold that is not tied to the median would also make PAvPU more informative for highly accurate models. Finally, since the uncertainty map is available before training, it could be used to guide targeted augmentation or resampling of high uncertainty and underrepresented regions, providing another way to address class imbalance beyond loss reweighting.

\section{Conclusion} \label{sec:conclusion}
This work investigated the use of geometric feature-based deterministic epistemic uncertainty in semantic segmentation in both 3D ALS and MLS point clouds through different roles. Modeling each class with a Gaussian distribution over the geometric descriptors provides an inexpensive, model independent and interpretable uncertainty estimate, with higher uncertainty generally occurring around thin, irregular and less represented structures.

Incorporating this uncertainty into the model in the role of, either an additional input feature or loss weighting, improves mIoU, helps recover minority classes and, in the majority of the configurations, improves the alignment between prediction errors and uncertainty. The mIoU gain holds for both baseline ANN models and state-of-the-art models, namely, PointNet++ and RandLA-Net, as for the point-wise network, including the strongest architecture in our study, which indicates that the benefit is not specific to a single architecture or to models of limited capacity. These results suggest that uncertainty can provide useful information beyond overall accuracy, particularly for identifying regions where segmentation predictions are less reliable. Our work provides the direction of incorporating uncertainties in the semantic segmentation models.

\bibliographystyle{unsrt}
\bibliography{papers_uncertainty}

@article{winkens2020contrastive,
  title={{Contrastive Training for Improved Out-of-Distribution Detection}},
  author={Winkens, Jim and Bunel, Rudy and Roy, Abhijit Guha and Stanforth, Robert and Natarajan, Vivek and Ledsam, Joseph R and MacWilliams, Patricia and Kohli, Pushmeet and Karthikesalingam, Alan and Kohl, Simon and Cemgil, Taylan and Eslami, S M Ali and Ronneberger, Olaf},
  journal={arXiv preprint arXiv:2007.05566},
  year={2020}
}

@inproceedings{postels2021hidden,
  title={{The Hidden Uncertainty in a Neural Network's Activations}},
  author={Postels, Janis and Blum, Hermann and Str{\"u}mpler, Yannick and Cadena, Cesar and Siegwart, Roland and Van Gool, Luc and Tombari, Federico},
  booktitle={ICML Workshop on Uncertainty \& Robustness in Deep Learning (UDL)},
  year={2021}
}

@inproceedings{postels2022practicality,
  title={{On the Practicality of Deterministic Epistemic Uncertainty}},
  author={Postels, Janis and Seg{\`u}, Mattia and Sun, Tao and Sieber, Luca Daniel and Van Gool, Luc and Yu, Fisher and Tombari, Federico},
  booktitle={Proceedings of the 39th International Conference on Machine Learning},
  series={Proceedings of Machine Learning Research},
  volume={162},
  pages={17870--17909},
  year={2022},
  publisher={PMLR}
}

@article{petschnigg2021uncertainty,
	author = {Petschnigg, Christina and Pilz, Jürgen},
	title = {{Uncertainty Estimation in Deep Neural Networks for Point Cloud Segmentation in Factory Planning}},
	journal = {Modelling},
	volume = {2},
	year = {2021},
	number = {1},
	pages = {1--17},
	url = {[https://www.mdpi.com/2673-3951/2/1/1](https://www.mdpi.com/2673-3951/2/1/1)},
	issn = {2673-3951},
	doi = {10.3390/modelling2010001}
}

@article{riedlinger2025,
    title={{LMD: Light-Weight Prediction Quality Estimation for Object Detection in Lidar Point Clouds}},
    author={Riedlinger, Tobias and Schubert, Marius and Penquitt, Sarina and Kezmann, Jan-Marcel and Colling, Pascal and Kahl, Karsten and Roese-Koerner, Lutz and Arnold, Michael and Zimmermann, Urs and Rottmann, Matthias},
    journal={{International Journal of Computer Vision}},
    volume={133},
    number={7},
    pages={4349--4365},
    year={2025},
    publisher={Springer},
    doi={10.1007/s11263-025-02377-8}
}

@inproceedings{yang2023geometry,
    author={Yang, Yuwei and Hayat, Munawar and Jin, Zhao and Ren, Chao and Lei, Yinjie},
    booktitle={{2023 IEEE/CVF Conference on Computer Vision and Pattern Recognition (CVPR)}}, 
    title={{Geometry and Uncertainty-Aware 3D Point Cloud Class-Incremental Semantic Segmentation}}, 
    year={2023},
    volume={},
    number={},
    pages={21759-21768},
    doi={10.1109/CVPR52729.2023.02084}
}

@article{mei2024improving,
    title={{Improving Deep Learning based Point Cloud Classification using Markov Random Fields with Quadratic Pseudo-Boolean Optimization}},
    author={Mei, Qipeng and Qiu, Kevin and Bulatov, Dimitri and Iwaszczuk, Dorota},
    journal={{ISPRS Annals of the Photogrammetry, Remote Sensing and Spatial Information Sciences}},
    volume={10},
    pages={229--236},
    year={2024},
    publisher={Copernicus GmbH},
    doi={10.5194/isprs-annals-X-4-W5-2024-229-2024}
}

@misc{li2025llm,
    title = {{LLM-Guided Taxonomy and Hierarchical Uncertainty for 3D Point Cloud Active Learning}}, 
    author = {Chenxi Li and Nuo Chen and Fengyun Tan and Yantong Chen and Bochun Yuan and Tianrui Li and Chongshou Li},
    year = {2025},
    eprint = {2505.18924},
    archivePrefix = {arXiv},
    primaryClass = {cs.CV},
    url = {https://arxiv.org/abs/2505.18924},
    doi = {10.48550/arXiv.2505.18924}
}

@inproceedings{xu2023hierarchical,
	author={Xu, Zongyi and Yuan, Bo and Zhao, Shanshan and Zhang, Qianni and Gao, Xinbo},
	booktitle={{2023 IEEE/CVF International Conference on Computer Vision (ICCV)}}, 
	title={{Hierarchical Point-based Active Learning for Semi-supervised Point Cloud Semantic Segmentation}}, 
	year={2023},
	volume={},
	number={},
	pages={18052-18062},
	doi={10.1109/ICCV51070.2023.01659}
}

@article{everingham2010pascal,
	title={{The Pascal Visual Object Classes (VOC) Challenge}},
	author={Everingham, Mark and Van Gool, Luc and Williams, Christopher KI and Winn, John and Zisserman, Andrew},
	journal={{International Journal of Computer Vision}},
	volume={88},
	number={2},
	pages={303--338},
	year={2010},
	publisher={Springer},
	doi={10.1007/s11263-009-0275-4}
}

@article{rottensteiner2012isprs,
	title={{The ISPRS benchmark on urban object classification and 3D building reconstruction}},
	author={Rottensteiner, Franz and Sohn, Gunho and Jung, Jaewook and Gerke, Markus and Baillard, Caroline and Benitez, Sebastien and Breitkopf, Uwe},
	journal={{ISPRS Annals of the Photogrammetry, Remote Sensing and Spatial Information Sciences}},
	volume={1},
	pages={293--298},
	year={2012},
	publisher={Copernicus GmbH},
	doi={10.5194/isprsannals-I-3-293-2012}
}

@article{roynard2018paris,
	author = {Roynard, Xavier and Deschaud, Jean-Emmanuel and Goulette, Fran{\c c}ois},
	title = {{Paris-Lille-3D: A large and high-quality ground-truth urban point cloud dataset for automatic segmentation and classification}},
	year = {2018},
	issue_date = {May 2018},
	publisher = {Sage Publications, Inc.},
	address = {USA},
	volume = {37},
	number = {6},
	issn = {0278-3649},
	url = {https://doi.org/10.1177/0278364918767506},
	doi = {10.1177/0278364918767506},
	journal = {Int. J. Rob. Res.},
	month = may,
	pages = {545–557},
	numpages = {13}
}

@inproceedings{tan2020toronto,
	author={Tan, Weikai and Qin, Nannan and Ma, Lingfei and Li, Ying and Du, Jing and Cai, Guorong and Yang, Ke and Li, Jonathan},
	booktitle={2020 IEEE/CVF Conference on Computer Vision and Pattern Recognition Workshops (CVPRW)}, 
	title={Toronto-3D: A Large-scale Mobile LiDAR Dataset for Semantic Segmentation of Urban Roadways}, 
	year={2020},
	volume={},
	number={},
	pages={797-806},
	doi={10.1109/CVPRW50498.2020.00109}
}

@misc{jakteristics,
    author={Caron, David},
    title={Jakteristics: Compute Geometric Features from Point Clouds},
    howpublished= {\url{https://jakteristics.readthedocs.io}},
    note={Accessed: 2026}
}

@inproceedings{lee2018mahalanobis,
    author = {Lee, Kimin and Lee, Kibok and Lee, Honglak and Shin, Jinwoo},
    booktitle = {{Advances in Neural Information Processing Systems}},
    editor = {S. Bengio and H. Wallach and H. Larochelle and K. Grauman and N. Cesa-Bianchi and R. Garnett},
    pages = {},
    publisher = {Curran Associates, Inc.},
    title = {{A Simple Unified Framework for Detecting Out-of-Distribution Samples and Adversarial Attacks}},
    url = {https://proceedings.neurips.cc/paper_files/paper/2018/file/abdeb6f575ac5c6676b747bca8d09cc2-Paper.pdf},
    volume = {31},
    year = {2018}
}

@inproceedings{mukhoti2023ddu,
    author={Mukhoti, Jishnu and Kirsch, Andreas and van Amersfoort, Joost and Torr, Philip H.S. and Gal, Yarin},
    booktitle={{2023 IEEE/CVF Conference on Computer Vision and Pattern Recognition (CVPR)}}, 
    title={{Deep Deterministic Uncertainty: A New Simple Baseline}}, 
    year={2023},
    volume={},
    number={},
    pages={24384-24394},
    doi={10.1109/CVPR52729.2023.02336}
}

@inproceedings{qi2017pointnetpp,
    author = {Qi, Charles Ruizhongtai and Yi, Li and Su, Hao and Guibas, Leonidas},
    booktitle = {{Advances in Neural Information Processing Systems}},
    editor = {I. Guyon and U. Von Luxburg and S. Bengio and H. Wallach and R. Fergus and S. Vishwanathan and R. Garnett},
    pages = {},
    publisher = {Curran Associates, Inc.},
    title = {{PointNet++: Deep Hierarchical Feature Learning on Point Sets in a Metric Space}},
    url = {https://proceedings.neurips.cc/paper_files/paper/2017/file/d8bf84be3800d12f74d8b05e9b89836f-Paper.pdf},
    volume = {30},
    year = {2017}
}

@misc{mukhoti2018pavpu,
    title={Evaluating Bayesian Deep Learning Methods for Semantic Segmentation}, 
    author={Jishnu Mukhoti and Yarin Gal},
    year={2019},
    eprint={1811.12709},
    archivePrefix={arXiv},
    primaryClass={cs.CV},
    url={https://arxiv.org/abs/1811.12709} 
}

@misc{ahn3,
    author={{Rijkswaterstaat and Het Waterschapshuis}},
    title={Actueel Hoogtebestand Nederland ({AHN3})},
    howpublished={\url{https://www.ahn.nl}},
    year={2019}
}

@inproceedings{kendall2017uncertainties,
    author = {Kendall, Alex and Gal, Yarin},
    title = {{What Uncertainties Do We Need in Bayesian Deep Learning for Computer Vision?}},
    year = {2017},
    isbn = {9781510860964},
    publisher = {{Curran Associates Inc.}},
    address = {Red Hook, NY, USA},
    pages = {5580–5590},
    numpages = {11},
    location = {Long Beach, California, USA},
    series = {NIPS'17},
    doi = {10.5555/3295222.3295309}
}

@article{roberts2017cv,
    title = {{Cross-Validation Strategies for Data with Temporal, Spatial, Hierarchical, or Phylogenetic Structure}},
    author = {Roberts, David R. and Bahn, Volker and Ciuti, Simone and Boyce, Mark S. and Elith, Jane and Guillera-Arroita, Gurutzeta and Hauenstein, Severin and Lahoz-Monfort, José J. and Schröder, Boris and Thuiller, Wilfried and Warton, David I. and Wintle, Brendan A. and Hartig, Florian and Dormann, Carsten F.},
    journal = {{Ecography}},
    volume = {40},
    number = {8},
    pages = {913-929},
    doi = {10.1111/ecog.02881},
    url = {https://nsojournals.onlinelibrary.wiley.com/doi/abs/10.1111/ecog.02881},
    eprint = {https://nsojournals.onlinelibrary.wiley.com/doi/pdf/10.1111/ecog.02881},
    year = {2017}
}

@article{ploton2020,
    title={{Spatial Validation Reveals Poor Predictive Performance of Large-Scale Ecological Mapping Models}},
    author={Ploton, Pierre and Mortier, Fr{\'e}d{\'e}ric and R{\'e}jou-M{\'e}chain, Maxime and Barbier, Nicolas and Picard, Nicolas and Rossi, Vivien and Dormann, Carsten and Cornu, Guillaume and Viennois, Ga{\"e}lle and Bayol, Nicolas and others},
    journal={{Nature Communications}},
    volume={11},
    number={1},
    pages={4540},
    year={2020},
    publisher={{Nature Publishing Group UK London}},
    doi={10.1038/s41467-020-18321-y}
}

@inproceedings{shojaei2025uncertainty,
    author={Shojaei Miandashti, Hanieh and Zou, Qianqian and Mehltretter, Max},
    editor={Del Bue, Alessio and Canton, Cristian and Pont-Tuset, Jordi and Tommasi, Tatiana},
    title={{Uncertainty Estimation and Out-of-Distribution Detection for LiDAR Scene Semantic Segmentation}},
    booktitle={Computer Vision -- ECCV 2024 Workshops},
    year={2025},
    publisher={{Springer Nature Switzerland}},
    address={Cham},
    pages={116--131},
    isbn={978-3-031-91767-7},
    doi={10.1007/978-3-031-91767-7_8}
}

@article{xu2025propagation,
    title = {{From Propagation to Prediction: Point-level Uncertainty Evaluation of MLS Point Clouds under Limited Ground Truth}},
    author = {Xu, Z. and Wysocki, O. and Holst, C.},
    journal = {ISPRS Annals of the Photogrammetry, Remote Sensing and Spatial Information Sciences},
    volume = {XI-1-2026},
    pages = {463--471},
    year = {2026},
    doi = {10.5194/isprs-annals-XI-1-2026-463-2026},
    url = {https://doi.org/10.5194/isprs-annals-XI-1-2026-463-2026}
}

@article{weinmann2014semantic,
    author = {Weinmann, M. and Jutzi, B. and Mallet, C.},
    title = {{Semantic 3D Scene Interpretation: A Framework Combining Optimal Neighborhood Size Selection with Relevant Features}},
    journal = {{ISPRS Annals of the Photogrammetry, Remote Sensing and Spatial Information Sciences}},
    volume = {II-3},
    pages = {181--188},
    year = {2014},
    doi = {10.5194/isprsannals-II-3-181-2014},
    url = {https://doi.org/10.5194/isprsannals-II-3-181-2014}
}

@article{weinmann2015semantic,
    title = {{Semantic Point Cloud Interpretation Based on Optimal Neighborhoods, Relevant Features and Efficient Classifiers}},
    journal = {{ISPRS Journal of Photogrammetry and Remote Sensing}},
    volume = {105},
    pages = {286-304},
    year = {2015},
    issn = {0924-2716},
    doi = {10.1016/j.isprsjprs.2015.01.016},
    url = {https://www.sciencedirect.com/science/article/pii/S0924271615000349},
    author = {Martin Weinmann and Boris Jutzi and Stefan Hinz and Clément Mallet}
}

@article{yao2024uncertainty,
	author={Yao, Baochen and Dong, Li and Qiu, Xiaojie and Song, Kangkang and Yan, Diqun and Peng, Chengbin},
	journal={{IEEE Transactions on Geoscience and Remote Sensing}}, 
	title={{Uncertainty-Guided Contrastive Learning for Weakly Supervised Point Cloud Segmentation}}, 
	year={2024},
	volume={62},
	number={},
	pages={1-13},
	doi={10.1109/TGRS.2024.3416219}
}

@article{qi2023nsa,
	author={Qi, Chao and Yin, Jianqin and Niu, Yingchun and Xu, Jinghang},
	journal={{IEEE Transactions on Geoscience and Remote Sensing}}, 
	title={{Neighborhood Spatial Aggregation MC Dropout for Efficient Uncertainty-Aware Semantic Segmentation in Point Clouds}}, 
	year={2023},
	volume={61},
	number={},
	pages={1-16},
	doi={10.1109/TGRS.2023.3314130}
}

@article{zhou2025dcuf,
	author={Zhou, Ce and Ling, Qiang},
	journal={IEEE Transactions on Geoscience and Remote Sensing}, 
	title={Weakly Supervised Large-Scale Point Cloud Semantic Segmentation Based on Dual Consistency Constraints and Uncertainty-Aware Fusion}, 
	year={2025},
	volume={63},
	number={},
	pages={1-13},
	doi={10.1109/TGRS.2024.3522060}
}

@article{shannon1948math,
	author={Shannon, C. E.},
	journal={{The Bell System Technical Journal}}, 
	title={{A mathematical theory of communication}}, 
	year={1948},
	volume={27},
	number={3},
	pages={379-423},
	doi={10.1002/j.1538-7305.1948.tb01338.x}
}

@article{hu2019randla,
  title={RandLA-Net: Efficient Semantic Segmentation of Large-Scale Point Clouds},
  author={Hu, Qingyong and Yang, Bo and Xie, Linhai and Rosa, Stefano and Guo, Yulan and Wang, Zhihua and Trigoni, Niki and Markham, Andrew},
  journal={Proceedings of the IEEE Conference on Computer Vision and Pattern Recognition},
  year={2020}
}

@article{hu2021learning,
  title={Learning Semantic Segmentation of Large-Scale Point Clouds with Random Sampling},
  author={Hu, Qingyong and Yang, Bo and Xie, Linhai and Rosa, Stefano and Guo, Yulan and Wang, Zhihua and Trigoni, Niki and Markham, Andrew},
  journal={IEEE Transactions on Pattern Analysis and Machine Intelligence},
  year={2021},
  publisher={IEEE}
}

@article{chen2024lfenet,
  author={Chen, Dong and Wang, Yuebin and Zhang, Liqiang and Kang, Zhizhong},
  journal={IEEE Transactions on Geoscience and Remote Sensing}, 
  title={Enhanced Local Feature Learning With Simple Offset Attention for Semantic Segmentation of Large-Scale Point Clouds}, 
  year={2024},
  volume={62},
  number={},
  pages={1-16},
  doi={10.1109/TGRS.2024.3453966}
}

@article{aditya2024vegetation,
  author={Aditya, Aditya and Lohani, Bharat and Aryal, Jagannath and Winter, Stephan},
  journal={IEEE Transactions on Geoscience and Remote Sensing}, 
  title={Benchmarking Deep Learning Architectures for Urban Vegetation Point Cloud Semantic Segmentation From MLS}, 
  year={2024},
  volume={62},
  number={},
  pages={1-14},
  doi={10.1109/TGRS.2024.3381976}
}

@article{jiang2022groundaware,
  author={Jiang, Tengping and Wang, Yongjun and Liu, Shan and Cong, Yangzi and Dai, Lei and Sun, Jian},
  journal={IEEE Transactions on Geoscience and Remote Sensing}, 
  title={Local and Global Structure for Urban ALS Point Cloud Semantic Segmentation With Ground-Aware Attention}, 
  year={2022},
  volume={60},
  number={},
  pages={1-15},
  doi={10.1109/TGRS.2022.3158362}
}

@article{guo2023mctnet,
  author={Guo, Bo and Deng, Liwei and Wang, Ruisheng and Guo, Wenchao and Ng, Alex Hay-Man and Bai, Wenfeng},
  journal={IEEE Transactions on Geoscience and Remote Sensing}, 
  title={MCTNet: Multiscale Cross-Attention-Based Transformer Network for Semantic Segmentation of Large-Scale Point Cloud}, 
  year={2023},
  volume={61},
  number={},
  pages={1-20},
  doi={10.1109/TGRS.2023.3322579}
}

@inproceedings{wu2024point,
  author={Wu, Xiaoyang and Jiang, Li and Wang, Peng-Shuai and Liu, Zhijian and Liu, Xihui and Qiao, Yu and Ouyang, Wanli and He, Tong and Zhao, Hengshuang},
  booktitle={2024 IEEE/CVF Conference on Computer Vision and Pattern Recognition (CVPR)}, 
  title={Point Transformer V3: Simpler, Faster, Stronger}, 
  year={2024},
  volume={},
  number={},
  pages={4840-4851},
  doi={10.1109/CVPR52733.2024.00463}
}

@inproceedings{lai2022stratified,
  author={Lai, Xin and Liu, Jianhui and Jiang, Li and Wang, Liwei and Zhao, Hengshuang and Liu, Shu and Qi, Xiaojuan and Jia, Jiaya},
  booktitle={2022 IEEE/CVF Conference on Computer Vision and Pattern Recognition (CVPR)}, 
  title={Stratified Transformer for 3D Point Cloud Segmentation}, 
  year={2022},
  volume={},
  number={},
  pages={8490-8499},
  doi={10.1109/CVPR52688.2022.00831}
}

@inproceedings{qian2022pointnext,
  author = {Qian, Guocheng and Li, Yuchen and Peng, Houwen and Mai, Jinjie and Al Kader Hammoud, Hasan Abed and Elhoseiny, Mohamed and Ghanem, Bernard},
  title = {PointNeXt: revisiting PointNet++ with improved training and scaling strategies},
  year = {2022},
  isbn = {9781713871088},
  publisher = {Curran Associates Inc.},
  address = {Red Hook, NY, USA},
  booktitle = {Proceedings of the 36th International Conference on Neural Information Processing Systems},
  articleno = {1685},
  numpages = {13},
  location = {New Orleans, LA, USA},
  series = {NIPS '22}
}

@article{puente2013mobile,
  title = {Review of Mobile Mapping and Surveying Technologies},
  journal = {Measurement},
  volume = {46},
  number = {7},
  pages = {2127-2145},
  year = {2013},
  issn = {0263-2241},
  doi = {https://doi.org/10.1016/j.measurement.2013.03.006},
  url = {https://www.sciencedirect.com/science/article/pii/S0263224113000730},
  author = {I. Puente and H. González-Jorge and J. Martínez-Sánchez and P. Arias}
}

@inproceedings{varney2020dales,
  title={{DALES: A Large-scale Aerial LiDAR Data Set for Semantic Segmentation}},
  author={Varney, Nina and Asari, Vijayan K and Graehling, Quinn},
  booktitle={2020 IEEE/CVF Conference on Computer Vision and Pattern Recognition Workshops (CVPRW)},
  year={2020},
  volume={},
  number={},
  pages={717-726},
  doi={10.1109/CVPRW50498.2020.00101}
}

@article{ye2020lasdu,
  AUTHOR = {Ye, Zhen and Xu, Yusheng and Huang, Rong and Tong, Xiaohua and Li, Xin and Liu, Xiangfeng and Luan, Kuifeng and Hoegner, Ludwig and Stilla, Uwe},
  TITLE = {LASDU: A Large-Scale Aerial LiDAR Dataset for Semantic Labeling in Dense Urban Areas},
  JOURNAL = {ISPRS International Journal of Geo-Information},
  VOLUME = {9},
  YEAR = {2020},
  NUMBER = {7},
  ARTICLE-NUMBER = {450},
  URL = {https://www.mdpi.com/2220-9964/9/7/450},
  ISSN = {2220-9964},
  DOI = {10.3390/ijgi9070450}
}

@article{hullermeier2021aleatoric,
  title={{Aleatoric and Epistemic Uncertainty in Machine Learning: An Introduction to Concepts and Methods}},
  author={H{\"u}llermeier, Eyke and Waegeman, Willem},
  journal={{Machine Learning}},
  volume={110},
  number={3},
  pages={457--506},
  year={2021},
  publisher={{Springer}},
  doi={10.1007/s10994-021-05946-3}
}

@article{araujo2024roadsafety,
  AUTHOR = {Araújo, Bernardo and Teixeira, João F. and Fonseca, Joaquim and Cerqueira, Ricardo and Beco, Sofia C.},
  TITLE = {The Road to Safety: A Review of Uncertainty and Applications to Autonomous Driving Perception},
  JOURNAL = {Entropy},
  VOLUME = {26},
  YEAR = {2024},
  NUMBER = {8},
  ARTICLE-NUMBER = {634},
  URL = {https://www.mdpi.com/1099-4300/26/8/634},
  PubMedID = {39202104},
  ISSN = {1099-4300},
  DOI = {10.3390/e26080634}
}
\end{document}